\def\year{2023}\relax
\documentclass[letterpaper]{article} 
\usepackage{aaai22}  
\usepackage{times}  
\usepackage{helvet}  
\usepackage{courier}  
\usepackage[hyphens]{url}  
\usepackage{graphicx} 
\usepackage{natbib}  
\usepackage{caption} 
\DeclareCaptionStyle{ruled}{labelfont=normalfont,labelsep=colon,strut=off} 
\usepackage{algorithm}
\usepackage{times}
\usepackage{latexsym}
\usepackage{booktabs}
\usepackage{url}
\usepackage{amssymb}
\usepackage{amsmath, calc}
\usepackage{multirow}
\usepackage{graphicx}
\usepackage{multirow}
\usepackage{makecell}
\usepackage{setspace}
\usepackage[noend]{algcompatible}
\usepackage{caption}
\usepackage{bm}
\usepackage{color}
\usepackage{subfig}

\usepackage{newfloat}
\usepackage{listings}
\floatstyle{ruled}
\newfloat{listing}{tb}{lst}{}
\floatname{listing}{Listing}
\title{Visually Grounded Commonsense Knowledge Acquisition}
\author{
    Yuan Yao$^{1}$, Tianyu Yu$^{2}$, Ao Zhang$^{4}$, Mengdi Li$^{5}$, Ruobing Xie$^{6}$, Cornelius Weber$^{4}$, \\ Zhiyuan Liu$^{1}$\thanks{Corresponding authors: Z.Liu (liuzy@tsinghua.edu.cn), H.Zh-eng (zheng.haitao@sz.tsinghua.edu.cn)}, Hai-Tao Zheng$^{2,3*}$, Stefan Wermter$^{5}$, Tat-Seng Chua$^{4}$, Maosong Sun$^{1}$
}
\affiliations{
    $^{1}$Dept. of Comp. Sci. \& Tech., Institute for AI, Tsinghua University, Beijing, China\\
    $^{2}$Shenzhen International Graduate School, Tsinghua University \\
    $^{3}$Peng Cheng Laboratory\\
    $^{4}$School of Computing, National University of Singapore, Singapore\\
    $^{5}$Department of Informatics, University of Hamburg, Hamburg, Germany \\
    $^{6}$WeChat AI, Tencent \\
    yaoyuanthu@163.com
}

\usepackage{bibentry}

\begin{document}

\maketitle

\begin{abstract}
Large-scale commonsense knowledge bases empower a broad range of AI applications, where the automatic extraction of commonsense knowledge (CKE) is a fundamental and challenging problem. CKE from text is known for suffering from the inherent sparsity and reporting bias of commonsense in text. Visual perception, on the other hand, contains rich commonsense knowledge about real-world entities, e.g., (\textit{person}, \texttt{can\_hold}, \textit{bottle}), which can serve as promising sources for acquiring grounded commonsense knowledge. In this work, we present CLEVER, which formulates \underline{C}KE as a distant\underline{L}y sup\underline{E}r\underline{V}ised multi-instanc\underline{E} lea\underline{R}ning problem, where models learn to summarize commonsense relations from a bag of images about an entity pair without any human annotation on image instances. To address the problem, CLEVER leverages vision-language pre-training models for deep understanding of each image in the bag, and selects 
informative instances from the bag to summarize commonsense entity relations via a novel contrastive attention mechanism. Comprehensive experimental results in held-out and human evaluation show that CLEVER can extract commonsense knowledge in promising quality, outperforming pre-trained language model-based methods by 3.9 AUC and 6.4 mAUC points. The predicted commonsense scores show strong correlation with human judgment with a 0.78 Spearman coefficient. Moreover, the extracted commonsense can also be grounded into images with reasonable interpretability. The data and codes can be obtained at \url{https://github.com/thunlp/CLEVER}.
\end{abstract}

\section{Introduction}

\begin{figure}[t]
    \centering
    \includegraphics[width=\columnwidth]{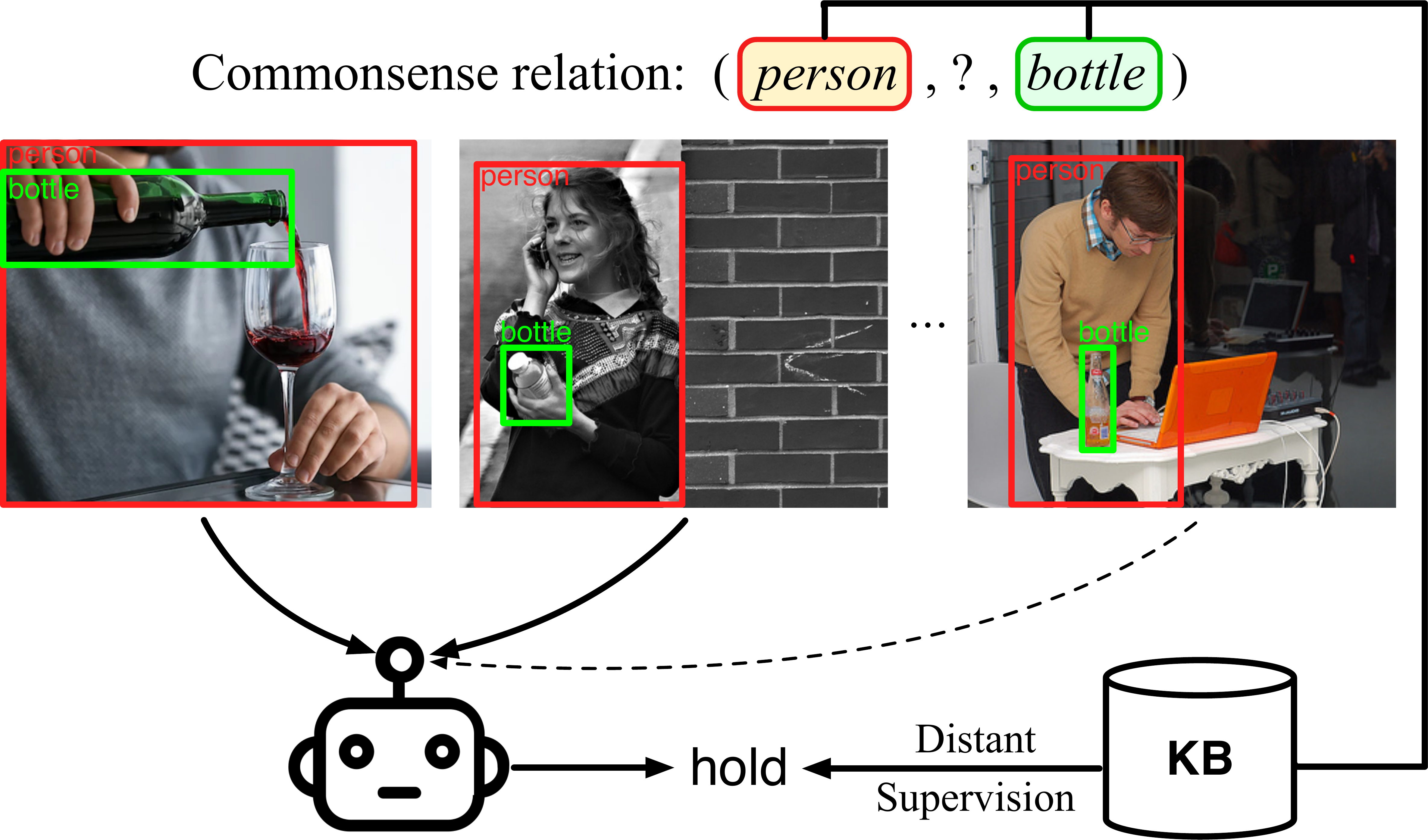}
    \caption{Visually grounded commonsense knowledge acquisition as a distantly supervised multi-instance learning problem. Given an entity pair and associated images, our model first understands entity interactions in each image, and then selects informative ones (solid line) to summarize the commonsense relations.}
    \label{fig:example}
\end{figure}

Providing machines with commonsense knowledge is a longstanding goal of artificial intelligence~\cite{davis1993knowledge}. Tremendous efforts have been devoted to building commonsense knowledge bases (KBs)~\cite{liu2004conceptnet,speer2017conceptnet,sap2019atomic}, which have facilitated various important applications in both computer vision~\cite{wu2017image,narasimhan2018out,gu2019scene,garderes-etal-2020-conceptbert} and natural language processing~\cite{zhou2018commonsense,wu2020diverse,lv2020graph}. However, most commonsense KBs 
are manually curated, which greatly limits their coverage and scale.

This paper studies the fundamental and challenging problem of commonsense knowledge extraction (CKE), which aims to extract plausible commonsense interactions between entities, e.g., (\textit{person}, \texttt{can\_hold}, \textit{bottle}). Previous works have attempted to extract commonsense knowledge from plain text~\cite{li2016commonsense} or pre-trained language models (PLMs)~\cite{petroni2019language,bosselut2019comet}. However, there is a growing consensus that obvious commonsense is rarely reported in text~\cite{gordon2013reporting,paik2021world}, and commonsense in PLMs suffers from low consistency and significant reporting bias ~\cite{shwartz2020neural,zhou2020evaluating,elazar2021measuring}. There is also widespread doubt whether learning purely from the surface text forms can lead to real understanding of commonsense meanings~\cite{bender-koller-2020-climbing}.

Visual perceptions (e.g., images), on the other hand, contain rich commonsense knowledge about real-world entities that can be consistently grounded. According to our statistics, $83\%$ of the triplets in visual relation learning datasets cannot be found in ConceptNet,\footnote{We randomly sample 200 distinct relational triplets from Visual Genome~\cite{krishna2017visual} and manually verify if the triplet or its variations are included in the ConceptNet.} indicating a promising direction for CKE from image data. However, most existing image-based CKE methods are either confined to restricted interaction types (e.g., spatial or partonomy relations)~\cite{chen2013neil,collell2018acquiring,xu2018automatic} or require extensive human annotation~\cite{vedantam2015learning}. 

In this work, we present CLEVER, which formulates \underline{C}KE as a distant\underline{L}y sup\underline{E}r\underline{V}ised multi-instanc\underline{E} lea\underline{R}ning problem~\cite{dietterich1997solving}, where models learn to summarize general commonsense relations of an entity pair from a bag of images, as shown in Figure~\ref{fig:example}. The commonsense relation labels are automatically created by aligning relational facts in existing KBs to image bags to provide distantly supervised learning signals. In this way, commonsense learning can easily scale up in general domain without costly manual image annotation.

To extract commonsense facts about a pair of query entities, models need to first understand their semantic interactions in each image of the bag, and then select informative ones (i.e., images that express interactions of interest between query entities) to synthesize the commonsense relations. However, our pilot experiments show that existing multi-instance learning methods cannot serve the task well, due to the complexity of real-world commonsense relations. Therefore, we propose a dedicated framework that models image-level entity interactions via vision-language pre-training (VLP) models, and selects meaningful images to summarize bag-level commonsense relations via a novel contrastive attention mechanism.

Comprehensive experimental results in held-out and human evaluation show that CLEVER can extract commonsense knowledge in promising quality, outperforming PLM-based approaches by 3.9 AUC and 6.4 mAUC points. The predicted commonsense scores show strong correlation with human judgment, achieving 0.78 Spearman's rank correlation coefficient. Moreover, the extracted commonsense can also be grounded into images with reasonable interpretability. Compared with PLM-based methods that produce commonsense purely based on text surface forms in a black-box fashion, the interpretability of CLEVER can be leveraged to provide supporting evidence for commonsense knowledge in KBs, which can be useful for downstream applications.


Our contributions are summarized as fourfold: (1) We propose to formulate CKE as a distantly supervised multi-instance learning problem, which can easily scale up for commonsense relations in a general domain without manual image annotation. (2) We conduct extensive experiments on existing and adapted CKE methods from different data sources, showing their effectiveness and limitations. (3) We present a dedicated CKE framework that integrates VLP models with a novel contrastive attention mechanism to deal with complex commonsense relation learning. (4) We conduct comprehensive experiments which demonstrate the effectiveness of the proposed framework.

\section{Related Work}

\textbf{Knowledge Bases.} Large-scale knowledge bases (KBs) that store abundant structured human knowledge facilitate various AI applications. Many efforts have been devoted to building KBs of different knowledge types, including linguistic knowledge~\cite{miller-1994-wordnet}, world knowledge~\cite{bollacker2008freebase} and commonsense knowledge~\cite{liu2004conceptnet,speer2017conceptnet,sap2019atomic}. However, existing KBs are mainly constructed with human annotation, which greatly limits their coverage and scale.

\smallskip
\textbf{Commonsense Knowledge Acquisition.} To acquire commonsense knowledge, some works attempt to learn from internal structures of existing triplets~\cite{speer2008analogyspace,malaviya2020commonsense}. However, these models usually suffer from the data sparsity of existing KBs. A more promising direction is to extract the commonsense contained in external data, i.e., commonsense knowledge extraction (CKE). Previous efforts in CKE can be divided into three categories according to the knowledge sources, including text-based, PLM-based and image-based models. 

(1) \textit{Text-based methods.} Early works attempt to extract commonsense from text~\cite{angeli2013philosophers,li2016commonsense}. However, CKE from text endures inherent reporting bias~\cite{gordon2013reporting}, i.e., people rarely state the obvious commonsense facts in text, making text not an ideal commonsense knowledge source. (2) \textit{PLM-based methods.} Since PLMs learn certain commonsense knowledge during pre-training, they can be probed or fine-tuned to generate commonsense knowledge~\cite{petroni2019language,davison-etal-2019-commonsense,bosselut2019comet}. However, it has been found that the commonsense in PLMs suffers from both low consistency, where small changes in the query templates can lead to substantially different predictions~\cite{zhou2020evaluating,elazar2021measuring}, and significant bias where the commonsense predictions can greatly differ from human judgments~\cite{shwartz2020neural,paik2021world}. (3) \textit{Image-based methods.} Some works have explored CKE from images that contain rich grounded commonsense knowledge. \citet{chen2013neil} learn partonomy (i.e, \texttt{part\_of}) and taxonomy (i.e., \texttt{is\_a}) commonsense from images. \citet{yatskar-etal-2016-stating,xu2018automatic} extract spatial commonsense (e.g., \texttt{located\_near}). \citet{chao2015mining} learn unary affordance commonsense about entities. \citet{vedantam2015learning,10.1145/3477495.3531992} extract more general commonsense interactions based on human annotation. \citet{sadeghi2015viske} mine commonsense based on spatial consistency of entities. Different from previous works, we extract general type commonsense interactions between entities without human annotation or restricted assumptions about commonsense knowledge.

\smallskip
\textbf{Scene Graph Generation.} Understanding visual interactions between objects also lies in the interest of scene graph generation~\cite{krishna2017visual,lu2016visual,xu2017scene,tang2020unbiased,yao2021visual,yao2021cpt,zhang2022fine}. Different from CKE which aims to summarize global commonsense relations between entities from a bag of images, the goal of scene graph generation is to identify the local relation in a specific image. Moreover, scene graph models usually require large amounts of image annotations, whereas the proposed distantly supervised CKE framework does not need annotated images.

\smallskip
\textbf{World Knowledge Acquisition.} The extraction of factual world knowledge, e.g., (\textit{Bob Dylan}, \texttt{composer}, \textit{Blowin’ in the Wind}), is an important tool to supplement world knowledge bases. Most works in world knowledge acquisition focus on text as the knowledge source~\cite{nguyen2015relation,soares2019matching,wu-etal-2019-open,dong2020meta,chen2021cil,yao2019docred,yao2021codred,zhang-etal-2021-open}, with some attempts in multimodal world knowledge acquisition~\cite{wen2021resin}. To alleviate human annotation, \citet{mintz-etal-2009-distant} propose distant supervision that aligns KBs to text to create noisy relation labels. Following works focus on dealing with the noise in distant supervision under the multi-instance learning formulation \cite{riedel2010modeling,zeng2015distant,liu2018neural}. The most widely adopted method is the selective attention model~\cite{lin2016neural} which selects high-quality instances in the bag based on the attention mechanism. In comparison, we aim to extract commonsense knowledge from bag of images. We find in our experiment that existing multi-instance learning models cannot serve the complex commonsense learning well, and therefore we propose a dedicated approach for the task.

\section{Pilot Experiment and Analysis}
\label{sec:analysis}
To investigate the effectiveness and limitation of existing CKE methods, we first perform an empirical study of representative methods from different information sources, including text-based, PLM-based and image-based models.

\smallskip
\textbf{Problem Definition.} CKE aims to extract commonsense relational triplet $(s, r, o)$, which depicts plausible interactions $r\in \mathcal{R}$ between entities $(s, o)$. For example, (\textit{person}, \texttt{can\_hold}, \textit{bottle}) reflects the commonsense knowledge that a person can hold a bottle. A special \texttt{NA} relation is also included, indicating no relation between the entity pair.


\smallskip
\textbf{Benchmark Construction.}
We construct the CKE benchmark based on Visual Genome~\cite{krishna2017visual}, which contains relational triplets about entities from real-world image data. Specifically, we select distinct triplets with the top $100$ entity types and relation types. For automatic held-out evaluation~\cite{mintz-etal-2009-distant}, we split the triplets into disjoint training, validation and test sets. Each entity pair is associated with Visual Genome images that contain the entities. The training/validation/test data contains 13,780/1,166/3,496 commonsense facts, 6,443/678/1,964 entity pairs, and 55,911/5,224/13,722 images respectively. 

\smallskip
\textbf{Existing CKE Models.} We select representative CKE models for empirical study. (1) \textit{Text-based models.} We adopt RTP~\cite{schuster-etal-2015-generating}, a widely used triplet parser, which extracts commonsense triplets from captions based on dependency trees. We extract triplets from Conceptual Caption~\cite{sharma2018conceptual} containing 3M captions, and obtain the confidence of the global triplets according to their frequency in the caption data. (2) \textit{PLM-based models.} We adopt LAMA~\cite{petroni2019language} that probes knowledge in BERT by filling the prompting template containing the query entity pair and the masked relation (e.g., ``\textit{person} \texttt{[MASK]} \textit{bottle}'').\footnote{We also experimented with masking the entities, and find that masking relations achieves better performance.} Following~\citet{lin-etal-2020-birds}, we further fine-tune the model based on the same prompts using the triplets in the training set to better learn the commonsense knowledge. Following~\citet{peng-etal-2020-learning}, we also adopt a vanilla fine-tuned BERT model which predicts relations based on the entity names using \texttt{[CLS]} token. 


\smallskip
\textbf{Multi-instance Learning for Image-based CKE.} Intuitively, images are raw visual perceptions of rich real-world entity interactions, which can serve as a scalable and promising information source for CKE. However, most existing image-based CKE methods are either restricted in relation types, or require manual image annotation.

For general and scalable commonsense KB construction, it is desirable to extract general type commonsense knowledge from large-scale images without human annotation. To this end, we propose to formulate CKE as a \textit{multi-instance learning} problem~\cite{dietterich1997solving}, where commonsense relation $r$ between entities $(s, o)$ is summarized from a bag of images $\mathcal{B}_{(s, o)}=\{v_i\}_{i=1}^N$ containing the entity pair. Inspired by \citet{mintz-etal-2009-distant}, we align existing commonsense KBs to image bags to provide distantly supervised learning signals. Specifically, the image bag $\mathcal{B}_{(s, o)}$ is labeled with the relation $r$ between $(s, o)$ in the KB, assuming that at least a subset of images in the bag expresses the triplet $(s, r, o)$, and there might be some images in the bag that do not express the triplet. To extract the commonsense triplet, models need to first understand the entity interactions in each image of the bag, and then select the meaningful ones to synthesize the commonsense relations. 

We note some works exploring problems in similar formulation in world knowledge extraction from text. To investigate the effectiveness of existing multi-instance learning methods for image-based CKE, we adapt representative approaches that select and summarize bag of instances using average pooling~\cite{lin2016neural}, at-least-one strategy~\cite{zeng2015distant}, or attention mechanism~\cite{lin2016neural}. 

Specifically, given a triplet $(s, r, o)$, we first select a bag of images containing the query entity pair. In practice, the number of candidate images can be large (e.g., $\sim$1,000), while only a small portion reflects entity interactions. To compose a proper size of image bags, inspired by \citet{zellers2018neural}, we select images with top spatial overlaps (i.e., intersection over union in pixels) of query entities, which are more likely to exhibit interactions. The query entity pair in each image of the bag is encoded into feature representations $\{\boldsymbol{v}_i\}_{i=1}^N$ using an adapted Neural Motif~\cite{zellers2018neural} model, a widely used CNN-based entity pair encoder.

To obtain the bag representation $\boldsymbol{B}_{(s,o)}$, (1) \textit{average pooling} (AVG)~\cite{lin2016neural} computes the mean of instance representations: $\boldsymbol{B}_{(s,o)} = \frac{1}{N}\sum_{i=1}^N \boldsymbol{v}_i$; (2) \textit{at-least-one strategy} (ONE)~\cite{zeng2015distant} selects the most likely instance: $\boldsymbol{B}_{(s,o)} = \boldsymbol{v}_j$, where $v_j$ achieves the highest score on the golden relation $r^*$ of the given training triplet; (3) \textit{attention mechanism} (ATT)~\cite{lin2016neural} computes the weighted sum of instance representations: $\boldsymbol{B}_{(s,o)} = \sum_{i=1}^N \alpha_i\boldsymbol{v}_i$, where the attention weight is computed based on the golden relation query: $\alpha_i = \text{Softmax}_i ( \boldsymbol{v}_i^\top\boldsymbol{r}^*)$. The bag representation $\boldsymbol{B}_{(s,o)}$ is optimized towards the golden label $r^*$ via a softmax classifier. During inference, since the relation label is unknown, ONE and ATT enumerate relation queries for the corresponding relation prediction score.

In addition to multi-instance learning based approaches, we also adapt visual relation detection models for image-based CKE. To simulate a scalable scenario, we randomly select a moderate number (i.e., 100) of image-level annotations for each relation from Visual Genome, and train a Neural Motif~\cite{zellers2018neural} model to predict the relation between an entity pair in specific images. During inference, the relation score of a bag is obtained by max pooling over relation scores of all images in the bag.

\begin{figure}[t]
    \centering
    \includegraphics[width=\columnwidth]{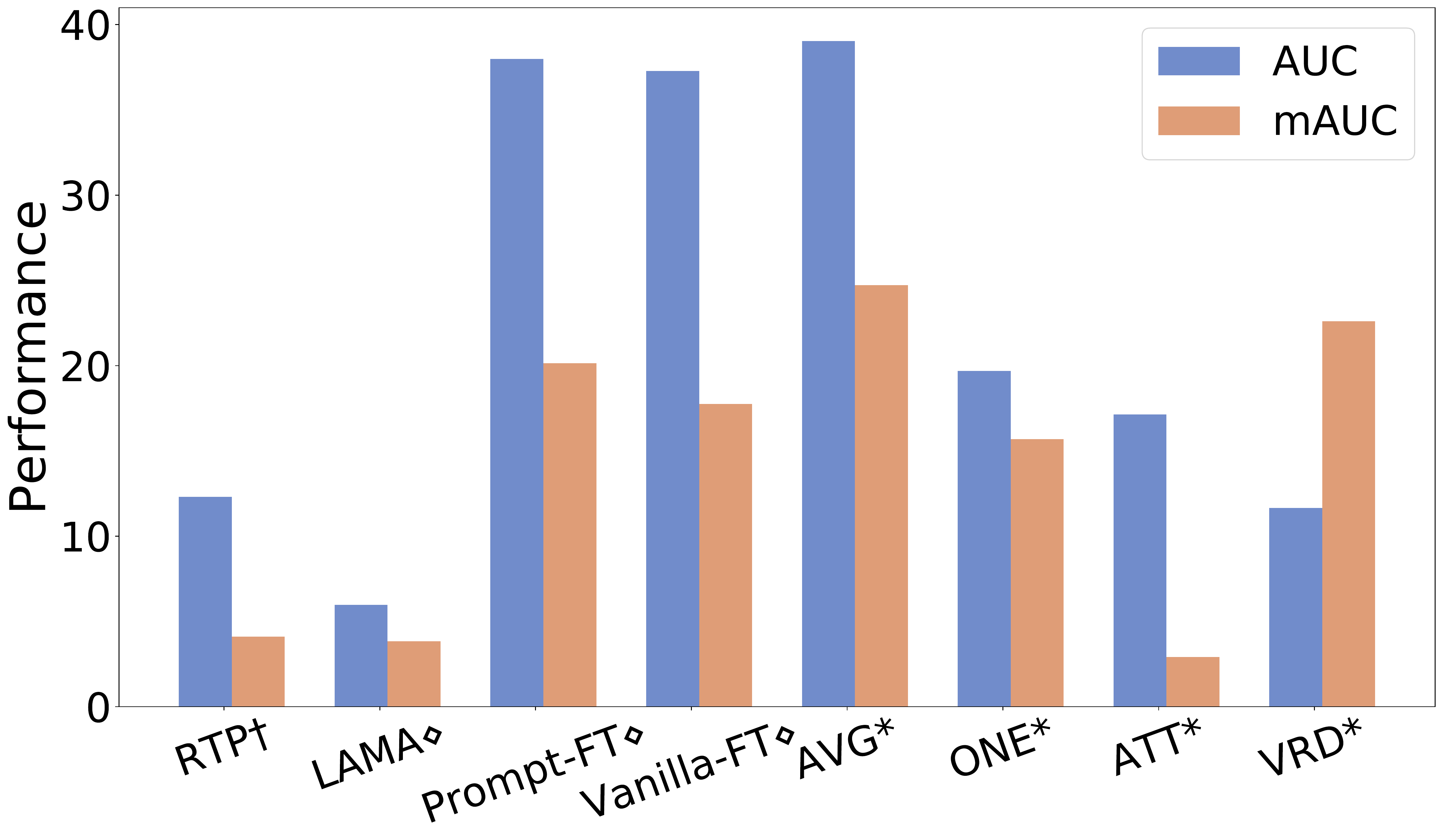}
    \caption{Results of CKE models from different information sources. $\dag$: text-based, $\Diamond$: PLM-based, *: image-based.}
    \label{fig:pilot experiment result}
\end{figure}

\smallskip
\textbf{Results.} Following previous works in knowledge acquisition~\cite{zeng2015distant,lin2016neural}, to provide a rigorous evaluation, we draw the precision-recall curve of held-out triplet predictions, and report the area under curve (AUC). Besides the traditional micro result, we also report mAUC, the area under the macro curve (i.e., the average curve of different relations) to evaluate the performance on long-tail relations. From Figure~\ref{fig:pilot experiment result} we have the following observations: 

(1) Text-based method (RTP) and knowledge probing from PLMs (LAMA) struggle on CKE. 
The reason is the inherent lack of commonsense knowledge in text, and the models are not fine-tuned for the task.
Further fine-tuning PLMs (Prompt-FT and Vanilla-FT) on the task can boost the performance to achieve a strong result. 

(2) Visual perceptions from images can provide rich information for commonsense knowledge acquisition. Based on a relatively proper summarization approach (AVG), multi-instance learning-based models on images achieve the best results over all existing CKE models. 

(3) Multi-instance learning formulation is necessary for scalable image-based CKE in open-domain. Adapted image-level visual relation detection models (VRD) do not perform well on CKE, despite more image-level relation annotations used (e.g., 100 image-level annotations per relation).

(4) Simple adaptation of existing multi-instance learning approaches cannot serve CKE well. The overall performance is still not satisfactory for all models. Notably, despite their competitive performance in world knowledge acquisition from text, ONE and ATT perform poorly on CKE. The reason is that compared with the relation schemes of world knowledge, commonsense relations exhibit higher complexity, where fine-grained relations with overlapping semantics (e.g., \texttt{stand\_on} and \texttt{walk\_on}), and hyponym-hypernym conflicts (e.g., \texttt{stand\_on} and \texttt{on}) frequently occur. Compared with AVG, the \textit{golden-query-only} problem of ONE and ATT hinders them from distinguishing complex commonsense relations. We refer readers to the methodology section for a more detailed discussion on the problem.

\begin{figure*}[t]
    \centering
    \hspace{10mm}\includegraphics[width=0.86\textwidth]{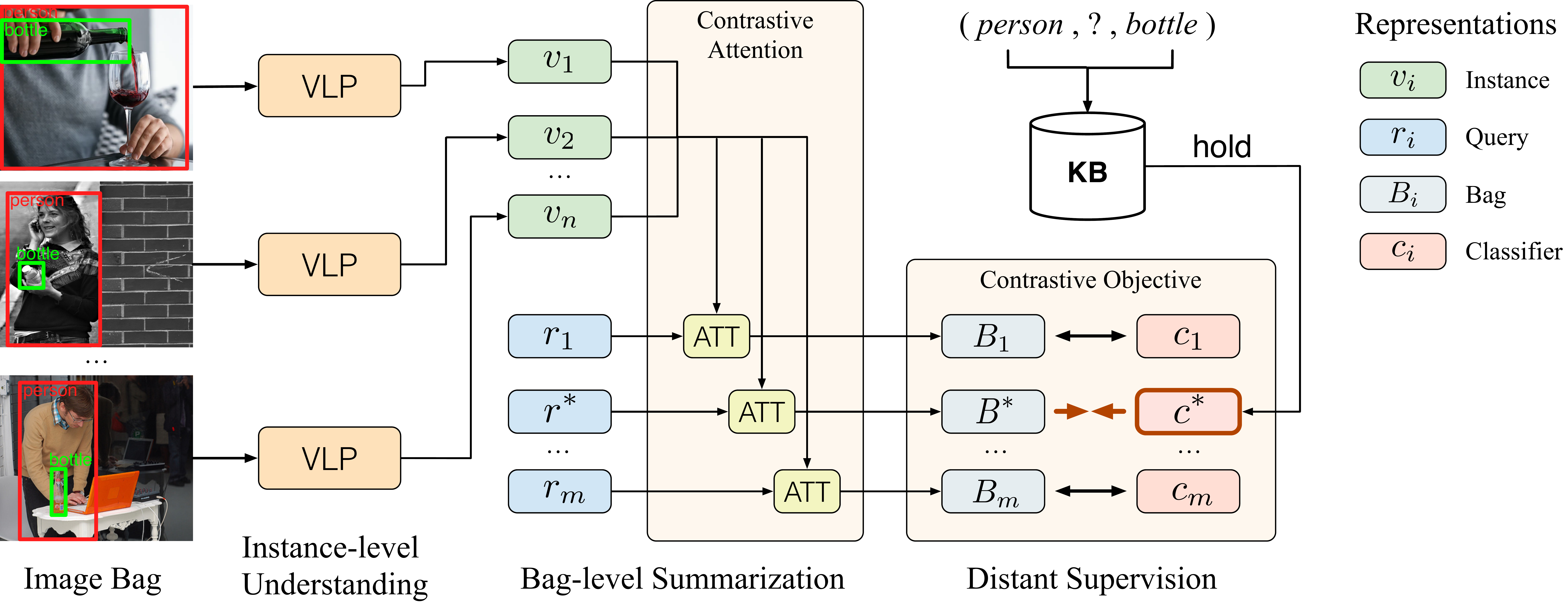}
    \caption{The CLEVER framework for visually grounded commonsense knowledge acquisition. Given a bag of images about an entity pair, our model leverages VLP models for image-level entity interaction understanding, and selects informative images to summarize bag-level commonsense relations via a contrastive attention mechanism.}
    \label{fig:framework}
\end{figure*}

\section{Methodology}

The pilot experiment results show that dedicated approaches need to be developed to address the unique challenges of commonsense knowledge acquisition. Essentially, due to the complexity of commonsense relations, multi-instance learning based CKE presents challenges on two levels: (1) on the image level, models need to first understand complex entity interactions in each image, (2) on the bag level, models are required to select informative instances to summarize the fine-grained commonsense relations between the entities. We present a dedicated model for CKE from images, as shown in Figure~\ref{fig:framework}, which (1) achieves deep understanding of the image-level interactions between entities through powerful vision-language pre-training (VLP) models, and (2) selects meaningful images to summarize bag-level commonsense relations via a contrastive attention mechanism.

\smallskip
\textbf{Vision-language Pre-training Models for Image-level Entity Interaction Understanding.} 
Recently VLP models have pushed forward the state-of-art of many multimodal tasks in a foundation role~\cite{bommasani2021opportunities}, such as visual question answering and visual grounding. However, few works have explored leveraging VLP methods to model complex visual relations for entity pairs. We show that pre-trained Transformers can serve as powerful foundation models to resolve complex image-level entity interactions.

Given a query entity pair $(s, o)$ and the associated image bag $\mathcal{B}_{(s, o)}=\{v_i\}_{i=1}^N$, each query entity pair instance in the bag is encoded into deep representations $\boldsymbol{v}_i$ via detector-based VLP models. In this work, we adopt VinVL~\cite{zhang2021vinvl}, a state-of-the-art VLP model as the encoder. Specifically, the query and context entities in each image are first encoded by object detectors to obtain a series of visual features $\{\boldsymbol{u}_1, \boldsymbol{u}_2, \dots, \boldsymbol{u}_n\}$. The visual features and token embeddings of entity tags $\{\boldsymbol{t}_1, \boldsymbol{t}_2, \dots, \boldsymbol{t}_n\}$ are then fed into pre-trained Transformers to obtain deep multimodal hidden representations $\{\boldsymbol{h}_u^1, \boldsymbol{h}_u^2, \dots, \boldsymbol{h}_u^n, \boldsymbol{h}_t^1, \boldsymbol{h}_t^2, \dots, \boldsymbol{h}_t^n\}$. The image-level entity pair representation is obtained by the concatenation of visual and text hidden representations: $\boldsymbol{v}_i=[\boldsymbol{h}_u^s; \boldsymbol{h}_u^o; \boldsymbol{h}_t^s; \boldsymbol{h}_t^o]$.

Despite the simplicity, the approach exhibits three important advantages in image-level entity interaction modeling: (1) The messages of entities (including query and context entities) are fused through multiple self-attention layers in Transformers to help model complex entity interactions. (2) Visual and textual information of entities are fused into deep multimodal representations. (3) Pre-trained deep vision-language representations are utilized to facilitate commonsense understanding. 

\smallskip
\textbf{Contrastive Attention Mechanism for Bag-level Commonsense Summarization.}
From the pilot experimental results, we observe that the complexity of commonsense relations (e.g., overlapping semantics and hyponym-hypernym conflicts) makes the relation boundaries hard to distinguish by existing multi-instance learning methods. In particular, despite its success in world knowledge acquisition from text, attention mechanism (ATT)~\cite{lin2016neural}  performs poorly on CKE. Here we identify that \textit{golden-query-only} is the key limitation of ATT in CKE, and show that by making the attention mechanism \textit{contrastive} over golden relation and other negative relations, the boundaries of complex commonsense relations can be effectively distinguished to achieve significantly better CKE performance.





We begin by discussing the \textit{golden-query-only} problem in ATT. During ATT training, the bag representation $\boldsymbol{B}_{(s, o)}$ is static for the prediction of different relations, and computed only based on the golden relation query. However, during inference, since the golden relation is unknown, all possible relations need to be enumerated to query the bag to predict the corresponding relation score. The golden-query-only problem leads to a lack of effective supervision for the bag representations (and relation scores) of other negative relations, resulting in indistinguishable negative bag representations from the golden ones.

To address the problem, we present a novel \textit{contrastive attention mechanism} that imposes contrastive supervision for golden and negative bag representations and relation scores. Specifically, for the prediction of \textit{each relation} $r_i\in \mathcal{R}$, a relation-aware bag representation $\boldsymbol{B}_{(s,r_i,o)}$ is obtained by a weighted sum of instance representations, where the attention weights are computed using the corresponding relation query $\boldsymbol{r}_i$ as follows:

\vspace{-0.5em}
{\small 
\begin{align}
    \boldsymbol{B}_{(s,r_i,o)} &= \sum_{j=1}^N \alpha^{r_i}_j\boldsymbol{v}_j, \\
    \alpha^{r_i}_j &= \text{Softmax}_j ( \boldsymbol{v}_j^\top\boldsymbol{r}_i).
\end{align}
}%

The bag representations are optimized via a contrastive InfoNCE loss~\cite{oord2018representation} as follows:

\begin{equation}
\small
\mathcal{L} = -\log \frac{\text{exp}(\boldsymbol{c}^{*\top}\boldsymbol{B}_{(s,r^*,o)})}{\sum_i\text{exp}( \boldsymbol{c}_i^\top\boldsymbol{B}_{(s,r_i,o)})},
\end{equation}
where $\boldsymbol{c}_i$ is the classifier embedding of $r_i$. In this way, the contrastive attention imposes clear boundaries between the bag representations of golden and negative relations to deal with the summarization of complex commonsense relations. The contrastive attention can also be viewed as a kind of cross-attention~\cite{vaswani2017attention} between relation queries and image instances, which can potentially benefit from multi-layer stacking. We leave it for future work.

\smallskip
\textbf{Integrating Multi-source Information for CKE.} Intuitively, multiple heterogeneous data sources can provide complementary information for commonsense learning. We show that this complementarity can be leveraged by a simple ensemble of models from each information source, where the aggregated triplet score is a weighted sum of the prediction score from each source. 

\section{Experiments}
\label{sec:experiment}
In this section, we empirically assess the effectiveness of the proposed model. We refer readers to the appendix for implementation details.

\begin{table*}[t]
    \centering
    \small
    \renewcommand\arraystretch{1.2}
    \scalebox{1.0}{
    \begin{tabular}{l|l|c c c|c c c}
    \toprule
    Source & Method & AUC  & F1 & P@2\% & mAUC  & mF1 & mP@2\% \\
    \midrule
    - & Random & \hspace{1.62mm}1.76 & \hspace{1.62mm}3.51 & \hspace{1.62mm}1.71 & \hspace{1.62mm}2.04 & \hspace{1.62mm}5.13 & \hspace{1.62mm}1.94 \\
    \midrule
    Text & RTP~\cite{schuster-etal-2015-generating} & 12.30 & 23.67 & 16.65 & \hspace{1.62mm}4.10 & \hspace{1.62mm}8.62 & \hspace{1.62mm}7.34 \\
    \midrule
    \multirow{3}{*}{PLM} & LAMA~\cite{petroni2019language} & \hspace{1.62mm}5.97 & 14.11 & 12.80 & \hspace{1.62mm}3.84 & \hspace{1.62mm}3.59 & \hspace{1.62mm}5.59 \\
    & Vanilla-FT~\cite{peng-etal-2020-learning} & 37.28 & 47.06 & 44.21 & 17.75 & 30.98 & 17.34 \\
    & Prompt-FT~\cite{lin-etal-2020-birds} & 37.99 & 44.43 & 41.69 & 20.15 & 35.37 & 19.81 \\
    \midrule
    \multirow{4}{*}{Image} & AVG~\cite{lin2016neural} & 39.04 & 47.49 & 44.34 & 24.73 & 41.07 & 20.83 \\
    & ONE~\cite{zeng2015distant} & 19.69 & 31.10 & 25.20 & 15.70 & 30.40 & 12.82   \\
    & ATT~\cite{lin2016neural} & 17.13 & 28.37 & 25.07 & \hspace{1.62mm}2.91 & \hspace{1.62mm}6.09 & \hspace{1.62mm}2.20 \\
    & CLEVER (Ours) & \underline{41.92} & \underline{48.96} & \underline{45.84} & \underline{26.57} & \textbf{43.62} & \underline{22.02} \\
    \midrule
    All & Ensemble (Ours)  & \textbf{45.68}  & \textbf{49.93} & \textbf{47.09} & \textbf{27.38} & \underline{43.13} & \textbf{22.80} \\

    \bottomrule
    \end{tabular}}
    \caption{Experimental results of CKE methods from different information sources. The best results are highlighted in bold, and best single model results are underlined.}
    \label{tab:main results}
\end{table*}

\begin{figure}[t]
    \centering
    \includegraphics[width=0.96\columnwidth]{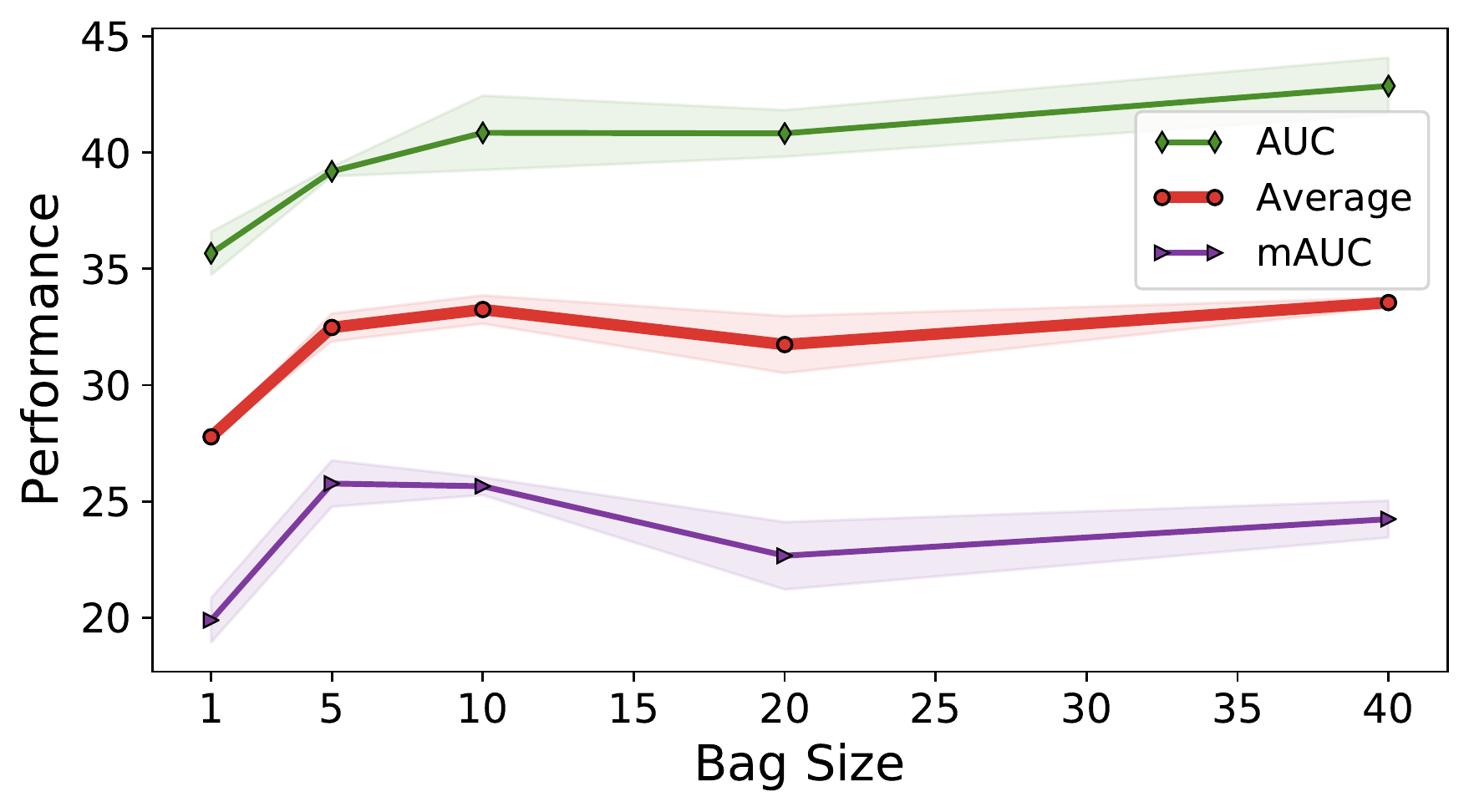}
    \vspace{-0.6em}
    \caption{Experimental results of our model with different bag sizes. We report AUC, mAUC and their average.}
    \label{fig:bag size}
    \vspace{-0.3em}
\end{figure}

\smallskip
\textbf{Experimental Settings.} (1) \textit{Benchmark and baselines.} We perform experiments on the CKE benchmark constructed from Visual Genome as described in the pilot experiment section, and compare to strong baselines from different information sources. We also include a random baseline that randomly predicts relations for entity pairs. For multi-source information integration, we ensemble CLEVER, RTP and Vanilla-FT. (2) \textit{Evaluation metrics.} To provide multi-dimensional evaluation, we also report the maximum F1 on curve, and precision@K\% (P@K\%) triplet prediction. 

\smallskip
\textbf{Main Results.} From the experimental results in Table~\ref{tab:main results}, we have the following observations: (1) CLEVER consistently achieves the best results among all baseline models in both micro and macro metrics. Specifically, CLEVER improves the performance of image-based models, and significantly outperforms previous best PLM-based results by 3.9 AUC and 6.4 mAUC points. The results show that CLEVER can extract commonsense knowledge from visual perceptions with promising quality. (2) Ensemble multi-source information further improves the performance over single-source models. This indicates that CKE can benefit from exploiting complementary information in different sources.

\begin{figure*}[t]
        \centering
        \subfloat[Informative Instances]{\includegraphics[width=0.46\textwidth]{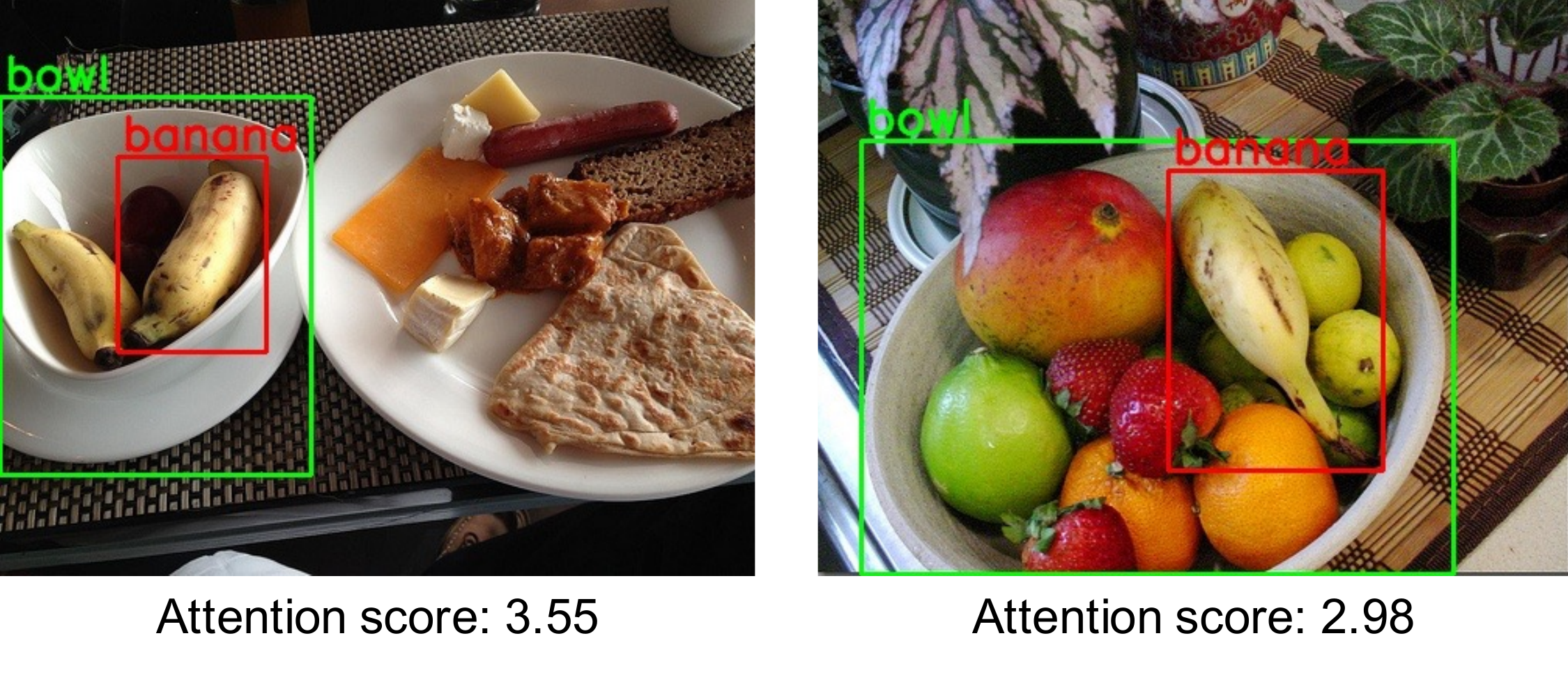}\label{fig:case1}}\hspace{5mm}
        \subfloat[Uninformative instances]{\includegraphics[width=0.46\textwidth]{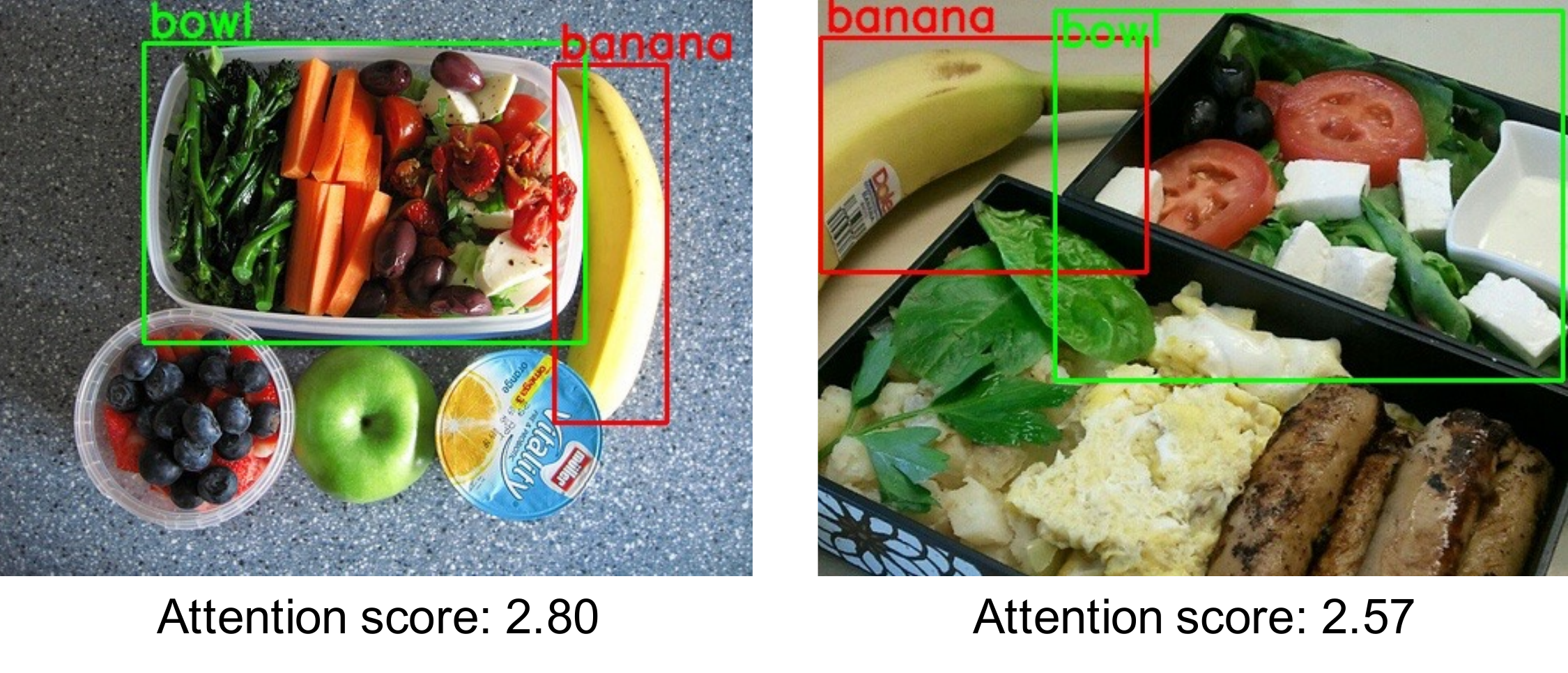}\label{fig:case2}}
        \vspace{-0.3em}
        \caption{Unnormalized attention scores of the extracted commonsense triplet (\textit{banana}, \texttt{in}, \textit{bowl}) over several images in a bag.}
        \label{fig:intepretability}
\end{figure*}
\vspace{-0.3em}

\begin{figure}[t]
    \centering
    \hspace{-1mm}
    \includegraphics[width=0.9\columnwidth]{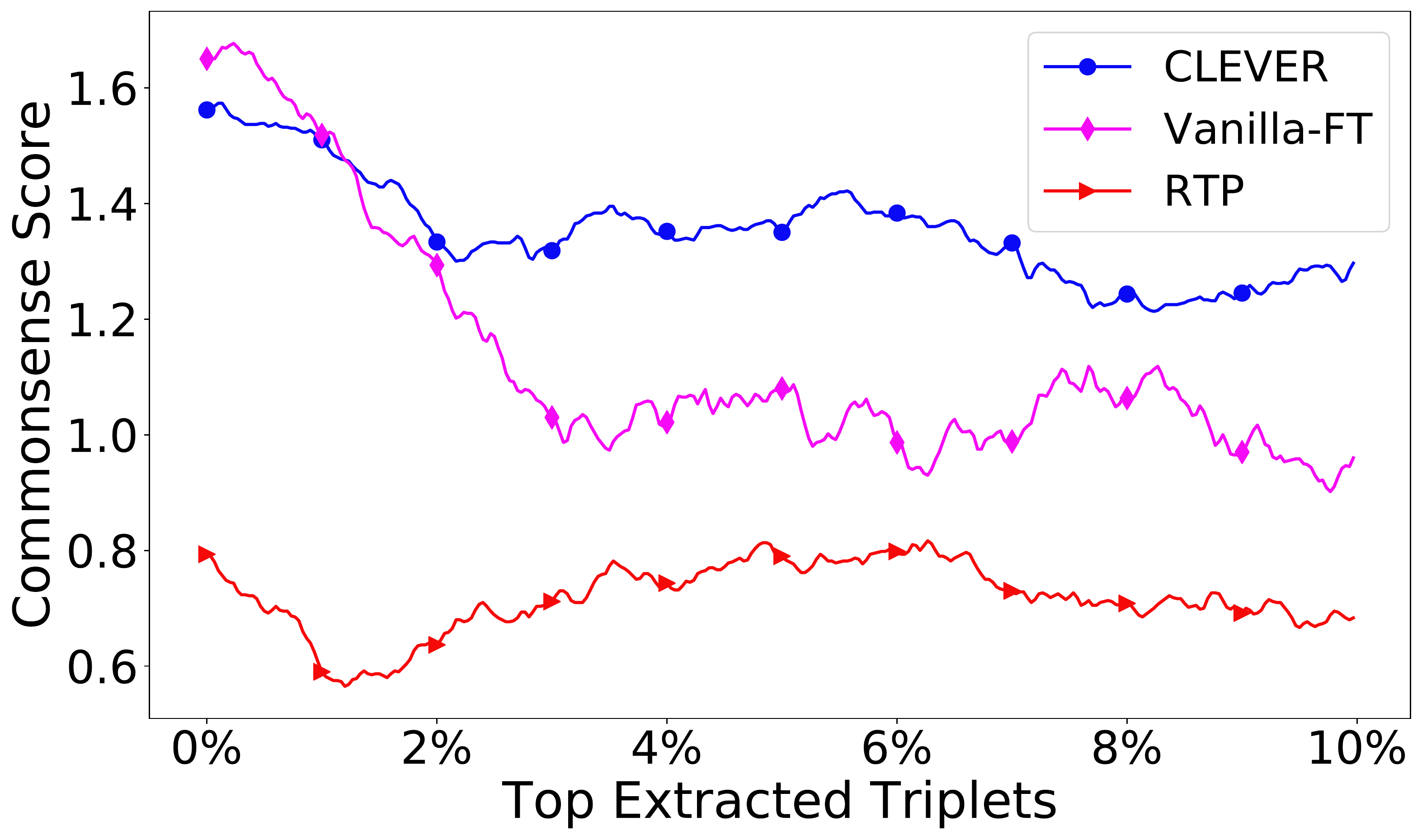}
    \caption{Human evaluation results on top extracted triplets.}
    \label{fig:human evaluation}
\end{figure}

\smallskip
\textbf{Human Evaluation.} In addition to the held-out evaluation, we also perform a human evaluation on top predictions. We select models that achieve the best micro performance on each source, including RTP, Vanilla-FT and CLEVER. Specifically, for each model, we sample from the top 10\% triplet predictions in a 1:50 ratio, resulting in 1,200 triplets for human evaluation. Each triplet is labeled by three independent annotators to decide the commonsense score: implausible (0), plausible but rare (1), common (2). We report the locally averaged triplet commonsense score given by human annotators in Figure~\ref{fig:human evaluation}. We can observe that triplets extracted by CLEVER are assigned with significantly higher commonsense scores in most cases. In addition, the commonsense scores of CLEVER achieve a strong 0.78 Spearman's rank correlation coefficient with human score, which shows that commonsense scores from our model can be well aligned to human judgments. The reason is that the contrastive attention mechanism can implicitly leverage the redundancy of instances to reflect the commonsense degree, where multiple informative instances in a bag can contribute to higher commonsense scores.

\begin{table}[t]
\begin{center}
\small
\begin{tabular}{l|cc|cc}
\toprule
Method & AUC & F1 & mAUC & mF1 \\
\midrule
CLEVER & \textbf{41.92} & \textbf{48.96} & \textbf{26.57} & \textbf{43.62} \\
\ \ \ \ VLP $\rightarrow$ CNN & 39.86 & 48.48 & 24.99 & 41.51 \\
\ \ \ \ CST-ATT $\rightarrow$ AVG & 39.95 & 47.73 & 25.56 & 41.51\\
\ \ \ \ CST-ATT $\rightarrow$ ONE &16.16 & 26.47& \hspace{1.62mm}5.17 & 13.00 \\
\ \ \ \ CST-ATT $\rightarrow$ ATT & 16.07  & 25.59 & \hspace{1.62mm}2.14 & \hspace{1.62mm}4.87 \\
\bottomrule
\end{tabular}
\end{center}
\caption{Ablation study on the instance encoder and the commonsense summarization method.}
\label{tab:ablation study}
\end{table}

\smallskip
\textbf{Interpretability.} 
In addition to the competitive performance, a crucial advantage of CLEVER is that the extracted commonsense knowledge can be grounded into visual perceptions through contrastive attention scores over image instances. As shown in Figure~\ref{fig:intepretability}, informative images are assigned with larger attention scores for commonsense learning. Compared with PLM-based approaches that produce commonsense knowledge purely based on correlations between text tokens in a black-box fashion, CLEVER enables trustworthy commonsense knowledge acquisition with better interpretability in the extraction process. From an application perspective, the selected informative images can also serve as supporting evidence for the extracted triplets in KBs for better knowledge utilization in downstream applications.

\smallskip
\textbf{Ablation Study.} We perform an ablation study by replacing the VLP encoder with the CNN-based encoder, and replacing the contrastive attention mechanism with existing multi-instance learning methods respectively. From the results in Table~\ref{tab:ablation study}, we can see that both components contribute to the final results. The results show that image-level entity interaction understanding and bag-level summarization are both important for good CKE performance.


\smallskip
\textbf{Effect of Bag Size.}
Intuitively, multiple images in a bag can provide diverse and complementary information about an entity pair for robust commonsense learning. To investigate the effect of bag size, we perform experiments on CLEVER with different bag sizes. From the results in Figure~\ref{fig:bag size}, we observe that: (1) A certain number of images is necessary to learn the commonsense interactions. The performance drops significantly when very small bag sizes are used. (2) The performance improvement is not significant when the bag size grows larger than $20$. We hypothesize the reason is that although a larger bag provides richer commonsense information, it also challenges the model with more noisy instances. Therefore, more advanced methods need to be developed to better exploit the rich information in larger image bags, which we leave for future work.

\smallskip
\textbf{Effect of Instance Sampling Strategy for Bag Construction.} Given the typically large number of open images containing an entity pair, it is desirable to select instances that are likely to express commonsense interactions at low costs to construct the bag. Besides the spatial overlap strategy, we experiment with another two sampling strategies: (1) \textit{Random sampling.} Random candidate images are selected to compose the bag. (2) \textit{CLIP-based sampling.} A text query is constructed for the entity pair as: ``\textit{s} has some relation with \textit{o}''. Then we encode the text query and image candidates using CLIP~\cite{radford2021learning}, and select the images with top similarity scores. We can see from Table~\ref{tab:bag sampling} that: (1) Entity interaction priors from CLIP and spatial overlap help select informative images for bag construction. (2) CLIP does not show significant advantage over spatial overlap. The reason is that spatial overlap incorporates more inductive bias for entity pair interactions, while CLIP is optimized to handle general sentences. Therefore, we choose spatial overlap for bag construction due to its simplicity and efficiency.

\begin{table}[t]
\begin{center}
\resizebox{\linewidth}{!}{%
\begin{tabular}{l|ccc|ccc}
\toprule
Method & AUC  & F1 & P@2\% & mAUC  & mF1 & mP@2\% \\
\midrule
Random & 41.3 & 45.7 & 43.0 & 23.7 & 38.0 & 21.7 \\
CLIP  & \textbf{44.2} & 47.3 & 44.5 & 24.0 & 38.5 & \textbf{22.6} \\
Overlap & 41.9 & \textbf{49.0} & \textbf{45.8} & \textbf{26.6} & \textbf{43.6} & 22.0 \\
\bottomrule
\end{tabular}}
\end{center}
\vspace{-0.5em}
\caption{Image sampling strategies for bag construction.}
\label{tab:bag sampling}
\end{table}

\begin{table}[!t]
    \definecolor{red}{RGB}{200,72,67}
    \centering
    \small
    \begin{tabular}[t]{cp{6.5cm}}
    \toprule
     Type & Examples  \\
    \midrule
     \multirow{3}{*}{\makecell[l]{ \multicolumn{1}{c}{\uppercase\expandafter{\romannumeral1}}}}  &  
     
     (\textit{woman}, \texttt{hold}, 	\textit{umbrella}), (\textit{horse}, \texttt{pull}, 	\textit{person}),
     
     (\textit{skateboard}, \texttt{under}, 	\textit{man}),  (\textit{flower}, \texttt{near}, 	\textit{fence}),

     (\textit{girl}, \texttt{wear}, 	\textit{glove}), (\textit{truck}, \texttt{has}, 	\textit{handle})	 	\\
     
     \cmidrule(lr{1em}){1-2}
\multirow{3}{*}{\makecell[l]{ \multicolumn{1}{c}{\uppercase\expandafter{\romannumeral2}}}}  &

     (\textit{snow}, \texttt{cover}, 	\textit{tire}), (\textit{cow}, \texttt{with}, 	\textit{nose}),

     (\textit{flower}, \texttt{in}, 	\textit{mountain}),  (\textit{wire}, \texttt{in}, 	\textit{building}),

     (\textit{logo}, \texttt{printed\_on}, 	\textit{train}), (\textit{boy}, \texttt{hold}, 	\textit{pillow})
     \\
     \cmidrule(lr{1em}){1-2}
\multirow{2}{*}{\makecell[l]{ \multicolumn{1}{c}{\uppercase\expandafter{\romannumeral3}}}}  &  
 (\textit{clock}, \texttt{has}, 	\textit{flower}), (\textit{boat}, \texttt{behind}, 	\textit{car}),

     (\textit{sheep}, \texttt{behind}, 	\textit{bench}),  (\textit{tail}, \texttt{on}, 	\textit{book}) \\
     
    \bottomrule
    \end{tabular}
    \caption{Extracted commonsense triplet examples in different types. \uppercase\expandafter{\romannumeral1}: Reasonable triplets unseen during training, \uppercase\expandafter{\romannumeral2}: novel facts for both Visual Genome and ConceptNet (i.e., newly discovered), \uppercase\expandafter{\romannumeral3}: Uncommonly observed facts.}
    \label{Table:case study}
\end{table}

\smallskip
\textbf{Case Study.} We provide examples of the extracted triplets from CLEVER in Table~\ref{Table:case study}. We can see that our model can extract reasonable commonsense knowledge unseen during training, and most importantly, novel facts to supplement commonsense KBs. We note that our model can sometimes produce uncommonly observed facts from accidental scene images. We refer readers to the appendix for the supporting images selected by our model for examples in type \uppercase\expandafter{\romannumeral3}.

\section{Conclusion}
In this work, we propose a novel formulation for commonsense knowledge acquisition as an image-based distantly supervised multi-instance learning problem. We present a dedicated framework that achieves deep image-level understanding via vision-language pre-training models, and bag-level summarization via a contrastive attention mechanism. Comprehensive experiments show the effectiveness of our framework. In the future, we will explore more advanced multi-instance learning approach, and acquire visual commonsense knowledge in more complex forms and types.

\section{Acknowledgements}
This work is funded by the Natural Science Foundation of China (NSFC 62061136001), the German Research Foundation (DFG TRR-169) in Project Crossmodal Learning,  National Natural Science Foundation of China (Grant No.62276154), AMiner.Shenzhen SciBrain Fund, Shenzhen Science and Technology Innovation Commission (Research Center for Computer Network (Shenzhen) Ministry of Education), Beijing Academy of Artificial Intelligence (BAAI), the Natural Science Foundation of Guangdong Province (Grant No. 2021A1515012640), Basic Research Fund of Shenzhen City (Grant No. JCYJ20210324120012033 and JSGG20210802154402007), and Overseas Cooperation Research Fund of Tsinghua Shenzhen International Graduate School (Grant No. HW2021008).

For author contributions, Yuan Yao designed the framework and experiments, and wrote the paper. Tianyu Yu conducted the experiments. Ao Zhang, Mengdi Li, Ruobing Xie, Cornelius Weber, Zhiyuan Liu, Hai-Tao Zheng, Stefan Wermter, Tat-Seng Chua and Maosong Sun provided valuable suggestions.

\bibliography{aaai22}

\begin{thebibliography}{70}
\providecommand{\natexlab}[1]{#1}

\bibitem[{Angeli and Manning(2013)}]{angeli2013philosophers}
Angeli, G.; and Manning, C.~D. 2013.
\newblock Philosophers are mortal: Inferring the truth of unseen facts.
\newblock In \emph{Proceedings of CoNLL}, 133--142.

\bibitem[{Bender and Koller(2020)}]{bender-koller-2020-climbing}
Bender, E.~M.; and Koller, A. 2020.
\newblock Climbing towards {NLU}: {On} Meaning, Form, and Understanding in the
  Age of Data.
\newblock In \emph{Proceedings of ACL}, 5185--5198.

\bibitem[{Bollacker et~al.(2008)Bollacker, Evans, Paritosh, Sturge, and
  Taylor}]{bollacker2008freebase}
Bollacker, K.; Evans, C.; Paritosh, P.; Sturge, T.; and Taylor, J. 2008.
\newblock Freebase: a collaboratively created graph database for structuring
  human knowledge.
\newblock In \emph{Proceedings of the ACM SIGMOD}, 1247--1250.

\bibitem[{Bommasani et~al.(2021)Bommasani, Hudson, Adeli, Altman, Arora, von
  Arx, Bernstein, Bohg, Bosselut, Brunskill
  et~al.}]{bommasani2021opportunities}
Bommasani, R.; Hudson, D.~A.; Adeli, E.; Altman, R.; Arora, S.; von Arx, S.;
  Bernstein, M.~S.; Bohg, J.; Bosselut, A.; Brunskill, E.; et~al. 2021.
\newblock On the opportunities and risks of foundation models.
\newblock \emph{arXiv preprint arXiv:2108.07258}.

\bibitem[{Bosselut et~al.(2019)Bosselut, Rashkin, Sap, Malaviya, Celikyilmaz,
  and Choi}]{bosselut2019comet}
Bosselut, A.; Rashkin, H.; Sap, M.; Malaviya, C.; Celikyilmaz, A.; and Choi, Y.
  2019.
\newblock COMET: Commonsense Transformers for Automatic Knowledge Graph
  Construction.
\newblock In \emph{Proceedings of ACL}, 4762--4779.

\bibitem[{Chao et~al.(2015)Chao, Wang, Mihalcea, and Deng}]{chao2015mining}
Chao, Y.-W.; Wang, Z.; Mihalcea, R.; and Deng, J. 2015.
\newblock Mining semantic affordances of visual object categories.
\newblock In \emph{Proceedings of ICCV}, 4259--4267.

\bibitem[{Chen et~al.(2021)Chen, Shi, Tang, Chen, Wu, and Zhuang}]{chen2021cil}
Chen, T.; Shi, H.; Tang, S.; Chen, Z.; Wu, F.; and Zhuang, Y. 2021.
\newblock CIL: Contrastive Instance Learning Framework for Distantly Supervised
  Relation Extraction.
\newblock In \emph{Proceedings of ACL}, 6191--6200.

\bibitem[{Chen, Shrivastava, and Gupta(2013)}]{chen2013neil}
Chen, X.; Shrivastava, A.; and Gupta, A. 2013.
\newblock NEIL: Extracting visual knowledge from web data.
\newblock In \emph{Proceedings of ICCV}, 1409--1416.

\bibitem[{Chen et~al.(2022)Chen, Zhang, Li, Deng, Tan, Xu, Huang, Si, and
  Chen}]{10.1145/3477495.3531992}
Chen, X.; Zhang, N.; Li, L.; Deng, S.; Tan, C.; Xu, C.; Huang, F.; Si, L.; and
  Chen, H. 2022.
\newblock Hybrid Transformer with Multi-Level Fusion for Multimodal Knowledge
  Graph Completion.
\newblock In \emph{Proceedings of ACM SIGIR}, 904–915.

\bibitem[{Collell, Van~Gool, and Moens(2018)}]{collell2018acquiring}
Collell, G.; Van~Gool, L.; and Moens, M.-F. 2018.
\newblock Acquiring common sense spatial knowledge through implicit spatial
  templates.
\newblock In \emph{Proceedings of AAAI}, volume~32.

\bibitem[{Davis, Shrobe, and Szolovits(1993)}]{davis1993knowledge}
Davis, R.; Shrobe, H.; and Szolovits, P. 1993.
\newblock What is a knowledge representation?
\newblock \emph{AI magazine}, 14(1): 17--17.

\bibitem[{Davison, Feldman, and Rush(2019)}]{davison-etal-2019-commonsense}
Davison, J.; Feldman, J.; and Rush, A. 2019.
\newblock Commonsense Knowledge Mining from Pretrained Models.
\newblock In \emph{Proceedings of EMNLP-IJCNLP}, 1173--1178.

\bibitem[{Dietterich, Lathrop, and
  Lozano-P{\'e}rez(1997)}]{dietterich1997solving}
Dietterich, T.~G.; Lathrop, R.~H.; and Lozano-P{\'e}rez, T. 1997.
\newblock Solving the multiple instance problem with axis-parallel rectangles.
\newblock \emph{Artificial intelligence}, 89(1-2): 31--71.

\bibitem[{Dong et~al.(2020)Dong, Yao, Xie, Gao, Han, Liu, Lin, Lin, and
  Sun}]{dong2020meta}
Dong, B.; Yao, Y.; Xie, R.; Gao, T.; Han, X.; Liu, Z.; Lin, F.; Lin, L.; and
  Sun, M. 2020.
\newblock Meta-information guided meta-learning for few-shot relation
  classification.
\newblock In \emph{Proceedings of the 28th International Conference on
  Computational Linguistics}, 1594--1605.

\bibitem[{Elazar et~al.(2021)Elazar, Kassner, Ravfogel, Ravichander, Hovy,
  Sch{\"u}tze, and Goldberg}]{elazar2021measuring}
Elazar, Y.; Kassner, N.; Ravfogel, S.; Ravichander, A.; Hovy, E.; Sch{\"u}tze,
  H.; and Goldberg, Y. 2021.
\newblock Measuring and improving consistency in pretrained language models.
\newblock \emph{TACL}, 9: 1012--1031.

\bibitem[{Gard{\`e}res et~al.(2020)Gard{\`e}res, Ziaeefard, Abeloos, and
  Lecue}]{garderes-etal-2020-conceptbert}
Gard{\`e}res, F.; Ziaeefard, M.; Abeloos, B.; and Lecue, F. 2020.
\newblock {C}oncept{B}ert: Concept-Aware Representation for Visual Question
  Answering.
\newblock In \emph{Findings of the Association for Computational Linguistics:
  EMNLP 2020}, 489--498.

\bibitem[{Gordon and Van~Durme(2013)}]{gordon2013reporting}
Gordon, J.; and Van~Durme, B. 2013.
\newblock Reporting bias and knowledge acquisition.
\newblock In \emph{Proceedings of the 2013 workshop on Automated knowledge base
  construction}, 25--30.

\bibitem[{Gu et~al.(2019)Gu, Zhao, Lin, Li, Cai, and Ling}]{gu2019scene}
Gu, J.; Zhao, H.; Lin, Z.; Li, S.; Cai, J.; and Ling, M. 2019.
\newblock Scene graph generation with external knowledge and image
  reconstruction.
\newblock In \emph{Proceedings of CVPR}, 1969--1978.

\bibitem[{Kingma and Ba(2014)}]{kingma2014adam}
Kingma, D.~P.; and Ba, J. 2014.
\newblock Adam: A method for stochastic optimization.
\newblock \emph{arXiv preprint arXiv:1412.6980}.

\bibitem[{Krishna et~al.(2017)Krishna, Zhu, Groth, Johnson, Hata, Kravitz,
  Chen, Kalantidis, Li, Shamma et~al.}]{krishna2017visual}
Krishna, R.; Zhu, Y.; Groth, O.; Johnson, J.; Hata, K.; Kravitz, J.; Chen, S.;
  Kalantidis, Y.; Li, L.-J.; Shamma, D.~A.; et~al. 2017.
\newblock Visual Genome: Connecting language and vision using crowdsourced
  dense image annotations.
\newblock \emph{IJCV}, 123(1): 32--73.

\bibitem[{Li et~al.(2016)Li, Taheri, Tu, and Gimpel}]{li2016commonsense}
Li, X.; Taheri, A.; Tu, L.; and Gimpel, K. 2016.
\newblock Commonsense knowledge base completion.
\newblock In \emph{Proceedings of ACL}, 1445--1455.

\bibitem[{Lin et~al.(2020)Lin, Lee, Khanna, and Ren}]{lin-etal-2020-birds}
Lin, B.~Y.; Lee, S.; Khanna, R.; and Ren, X. 2020.
\newblock {B}irds have four legs?! {N}umer{S}ense: {P}robing {N}umerical
  {C}ommonsense {K}nowledge of {P}re-{T}rained {L}anguage {M}odels.
\newblock In \emph{Proceedings of EMNLP}, 6862--6868.

\bibitem[{Lin et~al.(2016)Lin, Shen, Liu, Luan, and Sun}]{lin2016neural}
Lin, Y.; Shen, S.; Liu, Z.; Luan, H.; and Sun, M. 2016.
\newblock Neural relation extraction with selective attention over instances.
\newblock In \emph{Proceedings of ACL}, 2124--2133.

\bibitem[{Liu and Singh(2004)}]{liu2004conceptnet}
Liu, H.; and Singh, P. 2004.
\newblock ConceptNet—a practical commonsense reasoning tool-kit.
\newblock \emph{BT technology journal}, 22(4): 211--226.

\bibitem[{Liu et~al.(2018)Liu, Zhang, Zhou, and Jia}]{liu2018neural}
Liu, T.; Zhang, X.; Zhou, W.; and Jia, W. 2018.
\newblock Neural Relation Extraction via Inner-Sentence Noise Reduction and
  Transfer Learning.
\newblock In \emph{Proceedings of EMNLP}, 2195--2204.

\bibitem[{Lu et~al.(2016)Lu, Krishna, Bernstein, and Fei-Fei}]{lu2016visual}
Lu, C.; Krishna, R.; Bernstein, M.; and Fei-Fei, L. 2016.
\newblock Visual relationship detection with language priors.
\newblock In \emph{Proceedings of ECCV}, 852--869. Springer.

\bibitem[{Lv et~al.(2020)Lv, Guo, Xu, Tang, Duan, Gong, Shou, Jiang, Cao, and
  Hu}]{lv2020graph}
Lv, S.; Guo, D.; Xu, J.; Tang, D.; Duan, N.; Gong, M.; Shou, L.; Jiang, D.;
  Cao, G.; and Hu, S. 2020.
\newblock Graph-based reasoning over heterogeneous external knowledge for
  commonsense question answering.
\newblock In \emph{Proceedings of AAAI}, volume~34, 8449--8456.

\bibitem[{Malaviya et~al.(2020)Malaviya, Bhagavatula, Bosselut, and
  Choi}]{malaviya2020commonsense}
Malaviya, C.; Bhagavatula, C.; Bosselut, A.; and Choi, Y. 2020.
\newblock Commonsense knowledge base completion with structural and semantic
  context.
\newblock In \emph{Proceedings of AAAI}, volume~34, 2925--2933.

\bibitem[{Miller(1994)}]{miller-1994-wordnet}
Miller, G.~A. 1994.
\newblock {W}ord{N}et: A Lexical Database for {E}nglish.
\newblock In \emph{{H}uman {L}anguage {T}echnology: Proceedings of a Workshop
  held at {P}lainsboro, {N}ew {J}ersey, {M}arch 8-11, 1994}.

\bibitem[{Mintz et~al.(2009)Mintz, Bills, Snow, and
  Jurafsky}]{mintz-etal-2009-distant}
Mintz, M.; Bills, S.; Snow, R.; and Jurafsky, D. 2009.
\newblock Distant supervision for relation extraction without labeled data.
\newblock In \emph{Proceedings of ALC-IJCNLP}, 1003--1011.

\bibitem[{Moore(2013)}]{moore2013development}
Moore, C. 2013.
\newblock \emph{The development of commonsense psychology}.

\bibitem[{Narasimhan, Lazebnik, and Schwing(2018)}]{narasimhan2018out}
Narasimhan, M.; Lazebnik, S.; and Schwing, A. 2018.
\newblock Out of the box: Reasoning with graph convolution nets for factual
  visual question answering.
\newblock \emph{NeurIPS}, 31.

\bibitem[{Nguyen and Grishman(2015)}]{nguyen2015relation}
Nguyen, T.~H.; and Grishman, R. 2015.
\newblock Relation extraction: Perspective from convolutional neural networks.
\newblock In \emph{Proceedings of the 1st workshop on vector space modeling for
  natural language processing}, 39--48.

\bibitem[{Oord, Li, and Vinyals(2018)}]{oord2018representation}
Oord, A. v.~d.; Li, Y.; and Vinyals, O. 2018.
\newblock Representation learning with contrastive predictive coding.
\newblock \emph{arXiv preprint arXiv:1807.03748}.

\bibitem[{Paik et~al.(2021)Paik, Aroca-Ouellette, Roncone, and
  Kann}]{paik2021world}
Paik, C.; Aroca-Ouellette, S.; Roncone, A.; and Kann, K. 2021.
\newblock The World of an Octopus: How Reporting Bias Influences a Language
  Model’s Perception of Color.
\newblock In \emph{Proceedings of EMNLP}, 823--835.

\bibitem[{Peng et~al.(2020)Peng, Gao, Han, Lin, Li, Liu, Sun, and
  Zhou}]{peng-etal-2020-learning}
Peng, H.; Gao, T.; Han, X.; Lin, Y.; Li, P.; Liu, Z.; Sun, M.; and Zhou, J.
  2020.
\newblock {L}earning from {C}ontext or {N}ames? {A}n {E}mpirical {S}tudy on
  {N}eural {R}elation {E}xtraction.
\newblock In \emph{Proceedings of EMNLP}, 3661--3672.

\bibitem[{Pennington, Socher, and Manning(2014)}]{pennington2014glove}
Pennington, J.; Socher, R.; and Manning, C.~D. 2014.
\newblock Glove: Global vectors for word representation.
\newblock In \emph{Proceedings of the EMNLP}, 1532--1543.

\bibitem[{Petroni et~al.(2019)Petroni, Rockt{\"a}schel, Riedel, Lewis, Bakhtin,
  Wu, and Miller}]{petroni2019language}
Petroni, F.; Rockt{\"a}schel, T.; Riedel, S.; Lewis, P.; Bakhtin, A.; Wu, Y.;
  and Miller, A. 2019.
\newblock Language Models as Knowledge Bases?
\newblock In \emph{Proceedings of EMNLP-IJCNLP}, 2463--2473.

\bibitem[{Radford et~al.(2021)Radford, Kim, Hallacy, Ramesh, Goh, Agarwal,
  Sastry, Askell, Mishkin, Clark et~al.}]{radford2021learning}
Radford, A.; Kim, J.~W.; Hallacy, C.; Ramesh, A.; Goh, G.; Agarwal, S.; Sastry,
  G.; Askell, A.; Mishkin, P.; Clark, J.; et~al. 2021.
\newblock Learning transferable visual models from natural language
  supervision.
\newblock In \emph{ICML}, 8748--8763. PMLR.

\bibitem[{Ren et~al.(2015)Ren, He, Girshick, and Sun}]{ren2015faster}
Ren, S.; He, K.; Girshick, R.; and Sun, J. 2015.
\newblock Faster r-cnn: Towards real-time object detection with region proposal
  networks.
\newblock \emph{NeurIPS}, 28.

\bibitem[{Riedel, Yao, and McCallum(2010)}]{riedel2010modeling}
Riedel, S.; Yao, L.; and McCallum, A. 2010.
\newblock Modeling relations and their mentions without labeled text.
\newblock In \emph{Proceedings of ECML-PKDD}, 148--163. Springer.

\bibitem[{Sadeghi, Kumar~Divvala, and Farhadi(2015)}]{sadeghi2015viske}
Sadeghi, F.; Kumar~Divvala, S.~K.; and Farhadi, A. 2015.
\newblock Viske: Visual knowledge extraction and question answering by visual
  verification of relation phrases.
\newblock In \emph{Proceedings of ICCV}, 1456--1464.

\bibitem[{Sap et~al.(2019)Sap, Le~Bras, Allaway, Bhagavatula, Lourie, Rashkin,
  Roof, Smith, and Choi}]{sap2019atomic}
Sap, M.; Le~Bras, R.; Allaway, E.; Bhagavatula, C.; Lourie, N.; Rashkin, H.;
  Roof, B.; Smith, N.~A.; and Choi, Y. 2019.
\newblock Atomic: An atlas of machine commonsense for if-then reasoning.
\newblock In \emph{Proceedings of AAAI}, volume~33, 3027--3035.

\bibitem[{Schuster et~al.(2015)Schuster, Krishna, Chang, Fei-Fei, and
  Manning}]{schuster-etal-2015-generating}
Schuster, S.; Krishna, R.; Chang, A.; Fei-Fei, L.; and Manning, C.~D. 2015.
\newblock Generating Semantically Precise Scene Graphs from Textual
  Descriptions for Improved Image Retrieval.
\newblock In \emph{Proceedings of the Fourth Workshop on Vision and Language},
  70--80.

\bibitem[{Sharma et~al.(2018)Sharma, Ding, Goodman, and
  Soricut}]{sharma2018conceptual}
Sharma, P.; Ding, N.; Goodman, S.; and Soricut, R. 2018.
\newblock Conceptual captions: A cleaned, hypernymed, image alt-text dataset
  for automatic image captioning.
\newblock In \emph{Proceedings of ACL}, 2556--2565.

\bibitem[{Shwartz and Choi(2020)}]{shwartz2020neural}
Shwartz, V.; and Choi, Y. 2020.
\newblock Do neural language models overcome reporting bias?
\newblock In \emph{Proceedings of COLING}, 6863--6870.

\bibitem[{Soares et~al.(2019)Soares, Fitzgerald, Ling, and
  Kwiatkowski}]{soares2019matching}
Soares, L.~B.; Fitzgerald, N.; Ling, J.; and Kwiatkowski, T. 2019.
\newblock Matching the Blanks: Distributional Similarity for Relation Learning.
\newblock In \emph{Proceedings of ACL}, 2895--2905.

\bibitem[{Speer, Chin, and Havasi(2017)}]{speer2017conceptnet}
Speer, R.; Chin, J.; and Havasi, C. 2017.
\newblock Conceptnet 5.5: An open multilingual graph of general knowledge.
\newblock In \emph{Proceedings of AAAI}.

\bibitem[{Speer, Havasi, and Lieberman(2008)}]{speer2008analogyspace}
Speer, R.; Havasi, C.; and Lieberman, H. 2008.
\newblock AnalogySpace: Reducing the Dimensionality of Common Sense Knowledge.
\newblock In \emph{Proceedings of AAAI}, volume~8, 548--553.

\bibitem[{Tang et~al.(2020)Tang, Niu, Huang, Shi, and Zhang}]{tang2020unbiased}
Tang, K.; Niu, Y.; Huang, J.; Shi, J.; and Zhang, H. 2020.
\newblock Unbiased scene graph generation from biased training.
\newblock In \emph{Proceedings of CVPR}, 3716--3725.

\bibitem[{Vaswani et~al.(2017)Vaswani, Shazeer, Parmar, Uszkoreit, Jones,
  Gomez, Kaiser, and Polosukhin}]{vaswani2017attention}
Vaswani, A.; Shazeer, N.; Parmar, N.; Uszkoreit, J.; Jones, L.; Gomez, A.~N.;
  Kaiser, {\L}.; and Polosukhin, I. 2017.
\newblock Attention is all you need.
\newblock \emph{NeurIPS}, 30.

\bibitem[{Vedantam et~al.(2015)Vedantam, Lin, Batra, Zitnick, and
  Parikh}]{vedantam2015learning}
Vedantam, R.; Lin, X.; Batra, T.; Zitnick, C.~L.; and Parikh, D. 2015.
\newblock Learning common sense through visual abstraction.
\newblock In \emph{Proceedings of ICCV}, 2542--2550.

\bibitem[{Wen et~al.(2021)Wen, Lin, Lai, Pan, Li, Lin, Zhou, Li, Wang, Zhang
  et~al.}]{wen2021resin}
Wen, H.; Lin, Y.; Lai, T.; Pan, X.; Li, S.; Lin, X.; Zhou, B.; Li, M.; Wang,
  H.; Zhang, H.; et~al. 2021.
\newblock Resin: A dockerized schema-guided cross-document cross-lingual
  cross-media information extraction and event tracking system.
\newblock In \emph{Proceedings of NAACL}, 133--143.

\bibitem[{Wu et~al.(2017)Wu, Shen, Wang, Dick, and Van
  Den~Hengel}]{wu2017image}
Wu, Q.; Shen, C.; Wang, P.; Dick, A.; and Van Den~Hengel, A. 2017.
\newblock Image captioning and visual question answering based on attributes
  and external knowledge.
\newblock \emph{TPAMI}, 40(6): 1367--1381.

\bibitem[{Wu et~al.(2019)Wu, Yao, Han, Xie, Liu, Lin, Lin, and
  Sun}]{wu-etal-2019-open}
Wu, R.; Yao, Y.; Han, X.; Xie, R.; Liu, Z.; Lin, F.; Lin, L.; and Sun, M. 2019.
\newblock Open Relation Extraction: Relational Knowledge Transfer from
  Supervised Data to Unsupervised Data.
\newblock In \emph{Proceedings of EMNLP}, 219--228.

\bibitem[{Wu et~al.(2020)Wu, Li, Zhang, Zhou, and Wu}]{wu2020diverse}
Wu, S.; Li, Y.; Zhang, D.; Zhou, Y.; and Wu, Z. 2020.
\newblock Diverse and informative dialogue generation with context-specific
  commonsense knowledge awareness.
\newblock In \emph{Proceedings of ACL}, 5811--5820.

\bibitem[{Xu et~al.(2017)Xu, Zhu, Choy, and Fei-Fei}]{xu2017scene}
Xu, D.; Zhu, Y.; Choy, C.~B.; and Fei-Fei, L. 2017.
\newblock Scene graph generation by iterative message passing.
\newblock In \emph{Proceedings of ICCV}, 5410--5419.

\bibitem[{Xu, Lin, and Zhu(2018)}]{xu2018automatic}
Xu, F.~F.; Lin, B.~Y.; and Zhu, K. 2018.
\newblock Automatic Extraction of Commonsense LocatedNear Knowledge.
\newblock In \emph{Proceedings of ACL}, 96--101.

\bibitem[{Yao et~al.(2021{\natexlab{a}})Yao, Du, Lin, Li, Liu, Zhou, and
  Sun}]{yao2021codred}
Yao, Y.; Du, J.; Lin, Y.; Li, P.; Liu, Z.; Zhou, J.; and Sun, M.
  2021{\natexlab{a}}.
\newblock {CodRED}: A Cross-Document Relation Extraction Dataset for Acquiring
  Knowledge in the Wild.
\newblock In \emph{Proceedings of EMNLP}, 4452--4472.

\bibitem[{Yao et~al.(2019)Yao, Ye, Li, Han, Lin, Liu, Liu, Huang, Zhou, and
  Sun}]{yao2019docred}
Yao, Y.; Ye, D.; Li, P.; Han, X.; Lin, Y.; Liu, Z.; Liu, Z.; Huang, L.; Zhou,
  J.; and Sun, M. 2019.
\newblock {DocRED}: A Large-Scale Document-Level Relation Extraction Dataset.
\newblock In \emph{Proceedings of the ACL}, 764--777.

\bibitem[{Yao et~al.(2021{\natexlab{b}})Yao, Zhang, Han, Li, Weber, Liu,
  Wermter, and Sun}]{yao2021visual}
Yao, Y.; Zhang, A.; Han, X.; Li, M.; Weber, C.; Liu, Z.; Wermter, S.; and Sun,
  M. 2021{\natexlab{b}}.
\newblock Visual distant supervision for scene graph generation.
\newblock In \emph{Proceedings of the ICCV}, 15816--15826.

\bibitem[{Yao et~al.(2021{\natexlab{c}})Yao, Zhang, Zhang, Liu, Chua, and
  Sun}]{yao2021cpt}
Yao, Y.; Zhang, A.; Zhang, Z.; Liu, Z.; Chua, T.-S.; and Sun, M.
  2021{\natexlab{c}}.
\newblock {CPT}: Colorful prompt tuning for pre-trained vision-language models.
\newblock \emph{arXiv preprint arXiv:2109.11797}.

\bibitem[{Yatskar, Ordonez, and Farhadi(2016)}]{yatskar-etal-2016-stating}
Yatskar, M.; Ordonez, V.; and Farhadi, A. 2016.
\newblock Stating the Obvious: Extracting Visual Common Sense Knowledge.
\newblock In \emph{Proceedings of NAACL}, 193--198.

\bibitem[{Zellers et~al.(2018)Zellers, Yatskar, Thomson, and
  Choi}]{zellers2018neural}
Zellers, R.; Yatskar, M.; Thomson, S.; and Choi, Y. 2018.
\newblock Neural motifs: Scene graph parsing with global context.
\newblock In \emph{Proceedings of ICCV}, 5831--5840.

\bibitem[{Zeng et~al.(2015)Zeng, Liu, Chen, and Zhao}]{zeng2015distant}
Zeng, D.; Liu, K.; Chen, Y.; and Zhao, J. 2015.
\newblock Distant supervision for relation extraction via piecewise
  convolutional neural networks.
\newblock In \emph{Proceedings of EMNLP}, 1753--1762.

\bibitem[{Zhang et~al.(2022)Zhang, Yao, Chen, Ji, Liu, Sun, and
  Chua}]{zhang2022fine}
Zhang, A.; Yao, Y.; Chen, Q.; Ji, W.; Liu, Z.; Sun, M.; and Chua, T.-S. 2022.
\newblock Fine-Grained Scene Graph Generation with Data Transfer.
\newblock In \emph{Proceedings of ECCV}.

\bibitem[{Zhang et~al.(2021{\natexlab{a}})Zhang, Yao, Xie, Han, Liu, Lin, Lin,
  and Sun}]{zhang-etal-2021-open}
Zhang, K.; Yao, Y.; Xie, R.; Han, X.; Liu, Z.; Lin, F.; Lin, L.; and Sun, M.
  2021{\natexlab{a}}.
\newblock Open Hierarchical Relation Extraction.
\newblock In \emph{Proceedings of NAACL}, 5682--5693.

\bibitem[{Zhang et~al.(2021{\natexlab{b}})Zhang, Li, Hu, Yang, Zhang, Wang,
  Choi, and Gao}]{zhang2021vinvl}
Zhang, P.; Li, X.; Hu, X.; Yang, J.; Zhang, L.; Wang, L.; Choi, Y.; and Gao, J.
  2021{\natexlab{b}}.
\newblock Vinvl: Revisiting visual representations in vision-language models.
\newblock In \emph{Proceedings of CVPR}, 5579--5588.

\bibitem[{Zhou et~al.(2018)Zhou, Young, Huang, Zhao, Xu, and
  Zhu}]{zhou2018commonsense}
Zhou, H.; Young, T.; Huang, M.; Zhao, H.; Xu, J.; and Zhu, X. 2018.
\newblock Commonsense knowledge aware conversation generation with graph
  attention.
\newblock In \emph{IJCAI}, 4623--4629.

\bibitem[{Zhou et~al.(2020)Zhou, Zhang, Cui, and Huang}]{zhou2020evaluating}
Zhou, X.; Zhang, Y.; Cui, L.; and Huang, D. 2020.
\newblock Evaluating commonsense in pre-trained language models.
\newblock In \emph{Proceedings of AAAI}, volume~34, 9733--9740.

\end{thebibliography}

\appendix

\section{Additional Experiments}
In this section, we provide additional experimental results, including additional examples for the interpretability of commonsense triplet extractions, and supporting evidence selected by our model for uncommon facts.

\begin{figure*}[t]
    \centering
    \includegraphics[width=\textwidth]{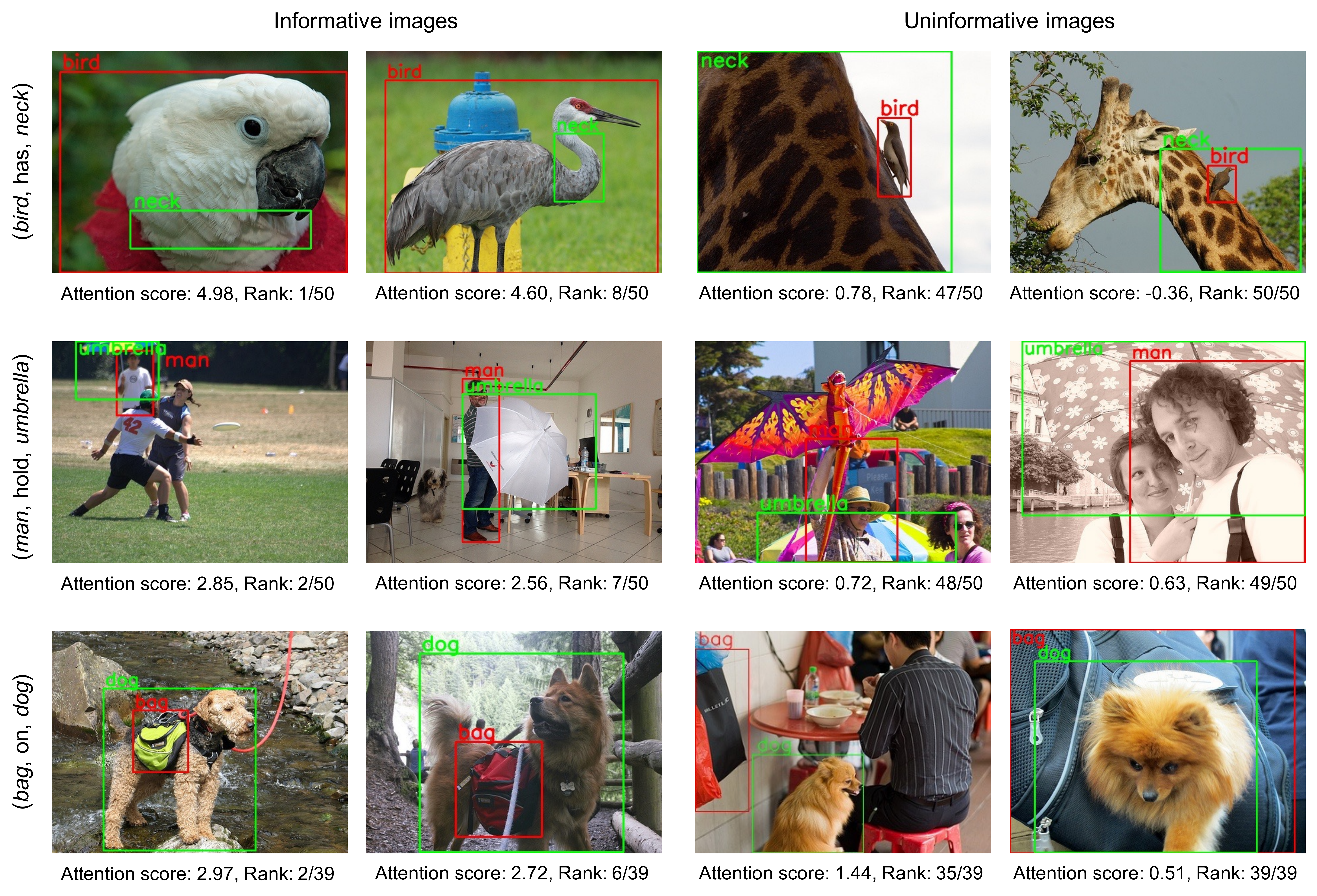}
    \caption{Unnormalized attention scores of the extracted commonsense triplets over images and their ranks in a bag.}
    \label{fig:interpretability}
\end{figure*}

\smallskip
\textbf{Additional Examples for Interpretability.} We provide more qualitative results for the interpretability of CLEVER commonsense extractions in Figure~\ref{fig:interpretability}. We can see that our model can extract reasonable commonsense knowledge involving diverse relations, including spatial relations, partonomy relations and actional relations. Moreover, the contrastive attention scores are discriminative over informative and uninformative images in the bag. This provides interpretability for the extraction process of CLEVER, making the resultant commonsense knowledge more trustworthy. Moreover, the selected informative images can serve as supporting evidence for the extracted commonsense knowledge in KBs, which can be useful for downstream applications.

\begin{figure*}[!t]
    \centering
    \includegraphics[width=\textwidth]{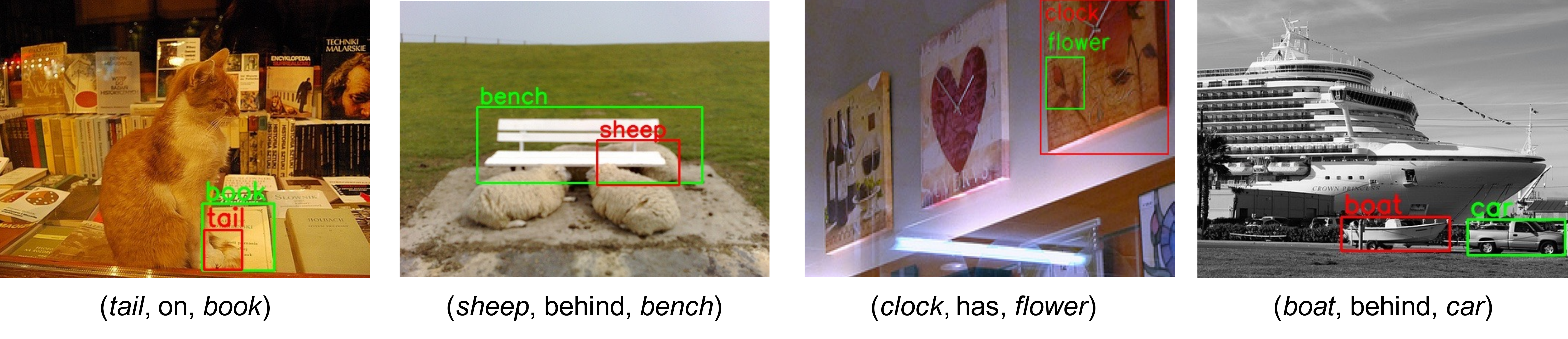}
    \vspace{-0.5em}
    \caption{Supporting images selected by our model for uncommonly observed facts.}
    \label{fig:uncommon_cases}
\end{figure*}

\smallskip
\textbf{Supporting Evidence for Uncommon Facts.} From the case study in the paper (type \uppercase\expandafter{\romannumeral3} in Table 4), we observe that our model can sometimes produce uncommonly observed facts from accidental scene images. We show the supporting images selected by our model with high attention scores in Figure~\ref{fig:uncommon_cases}. Although the facts are not common, their plausibility can be easily verified once the supporting images are provided. This demonstrates the advantage of extracting commonsense knowledge from grounded information sources, and the effectiveness of the proposed model in image-level entity interaction understanding and bag-level selection.

\section{Discussion and Outlook}
Despite the promising results of the proposed model, we note that there is still ample room for improvement. In this section, we discuss the limitation of this work and promising directions for future research.

\begin{table*}
    \begin{center}
    
    \small
    \begin{tabular}{ll ll ll ll ll ll}
    \toprule
man & person & woman & tree & building & table & sign & boy & window & fence & pole & girl \\
dog & snow & car & bench & street & train & bird & light & head & chair & hand & sidewalk \\
door & bike & elephant & rock & horse & bus & glass & truck & bag & box & boat & beach \\
plate & clock & leaf & plant & board & umbrella & giraffe & leg & flower & motorcycle & track & cow \\
post & hill & zebra & surfboard & banana & shirt & shelf & house & face & food & wire & arm \\
hair & skateboard & paper & branch & bottle & handle & sheep & roof & bowl & wheel & book & logo \\
trunk & cup & mountain & lamp & seat & shoe & wave & pillow & jacket & cabinet & hat & tail \\
letter & tire & ear & nose & helmet & eye & cap & coat & mouth & glove & pant & tile \\
neck & jean & short & wing \\
    \bottomrule
    \end{tabular}
    \end{center}
    \caption{List of entities of the benchmark.}
    \label{table:entities}
\end{table*}

\begin{table*}
    \begin{center}

    \small
    \begin{tabular}{ll ll ll ll ll}
    \toprule
on & near & in & has & behind & with & above & of \\
under & in front of & holding & sitting on & over & attached to & wearing & standing on \\
for & between & looking at & at & hanging from & standing next to & belonging to & covering \\
touching & underneath & carrying & next & laying on & outside & wears & part of \\
on back of & beneath & leaning on & along & riding & standing by & watching & standing behind \\
standing near & from & against & to & standing & resting on & riding on & across \\
sitting on top of & mounted on & walking on & lying on & worn by & covered in & has on & in middle of \\
sitting & walking & eating & painted on & connected to & holding up & using & covered with \\
surrounding & growing on & held by & crossing & walking in & made of & supporting & full of \\
pulling & lining & playing with & printed on & filled with & walking down & parked on & on bottom of \\
laying in & cutting & lying on top of & contains & sitting at & hitting & built into & shows \\
parked in & written on & playing & says & driving on & adorning & growing in & hanging in \\
swinging & flying & throwing & floating in \\
    \bottomrule
    \end{tabular}
    
    \end{center}
    \caption{List of relations of the  benchmark.}
    \label{table:relations}
\end{table*}

\smallskip
\textbf{Commonsense in More Complex Forms.}
In this work, we formulate commonsense as a triplet of binary commonsense relations between a pair of entities. Although binary relations constitute a fundamental part of real-world commonsense, humans can summarize commonsense from images in more complex forms: (1) \textit{N-ary commonsense interactions among multiple entities.} For example, the commonsense ``person can write letters with a pen'' can be summarized as a ternary relation among three entities   \texttt{write}(\textit{person}, \textit{letter}, \textit{pen}). (2) \textit{Commonsense correlations between structured facts.} For example, (\textit{rainwater}, \texttt{on}, \textit{road}) is highly correlated with (\textit{person}, \texttt{hold}, \textit{umbrella}).

\smallskip
\textbf{Commonsense in More Complex Types.}
Although images contain rich commonsense knowledge about the visual worlds, we note that there are still important commonsense types out of reach of images: (1) \textit{Temporal commonsense.} For example, a person needs to open the door of a refrigerator before getting the milk in it, which can hopefully be acquired from videos. (2) \textit{Invisible commonsense.} For example, (\textit{love}, \texttt{can cause}, \textit{happiness}) may arguably only be extracted from text. Although PLMs can deal with flexible forms and types of commonsense, there is a general belief that learning purely from the correlation of surface text forms without grounding to real-world perceptions cannot lead to real understanding of commonsense meanings~\cite{bender-koller-2020-climbing}. Moreover, a developmental learning procedure from concrete grounded commonsense (from visual perceptions) to abstract commonsense (from language) is also more bio-plausible and supported by human cognition~\cite{moore2013development}.

\section{Implementation Details}

In this section, we provide the implementation details of the experiments, including benchmark statistics, model training, CNN encoder and evaluation metrics.

\smallskip
\textbf{Benchmark Statistics.} To construct the benchmark, we select distinct triplets with the top 100 entity categories and relation categories. Here we provide the category list of entities and relations of the benchmark in Table~\ref{table:entities} and Table~\ref{table:relations}.

\smallskip
\textbf{Bag Construction.} For bag construction, the entity pairs in images are ranked according to their intersection over union in images. For entity pairs that do not have overlapping areas in an image, they are ranked according to the distance between the central points of the entities.

\smallskip
\textbf{Model Training.} The hyperprameters in the experiments are selected by grid search according to the average of AUC and mACU scores on the validation set. We use base-size pre-trained models in all our experiments. Our best model is trained using AdamW~\cite{kingma2014adam} optimizer on 10 NVIDIA GeForce RTX 2080 Ti for 18 epochs, with bag size 50, learning rate 7e-5, batch size 60 and weight decay 0.01. We first warm up the training by linearly increasing the learning rate from 7e-6 to 7e-5 in 1000 steps. The learning rate decreases by 10 times after the performance plateau on the validation set, and the training terminates after three performance plateaus. 

\smallskip
\textbf{CNN Encoder.} We adapt a Neural Motif~\cite{zellers2018neural} model to encode image-level entity pair features for AVG, ONE and ATT. Specifically, we adopt a Faster R-CNN \cite{ren2015faster} model pre-trained on Visual Genome as the visual encoder to extract raw features for objects. For each query entity pair, we concatenate two raw feature vectors and feed the result into an MLP with LayerNorm and ReLU activation to generate the visual representation. Following the implementation of \citet{zellers2018neural}, we also utilize pre-trained GloVe vectors \cite{pennington2014glove} of entity categories as extra information from text-domain. The Bi-LSTM context encoder is removed since the limited parallelization capability prevents the model from converging at an acceptable time cost. We also conduct few experiments to study the effectiveness of spatial features represented by bounding box positions. Models trained with the usage of spatial features achieve merely marginal performance gain with three times more memory occupation. Therefore, we do not include the spatial feature to the input of our CNN-based model.

\smallskip
\textbf{Evaluation Metrics.} Following previous works in knowledge acquisition from text~\cite{zeng2015distant,lin2016neural}, to provide a rigorous evaluation, we report results based on the precision-recall curve of held-out triplet predictions. Specifically, given a query entity pair $(s, o)$, our model predicts a commonsense score for each relation $r_i$ (excluding \texttt{NA}), which indicates the plausibility of a potentially useful triplet $(s, r_i, o)$. We rank all candidate triplets according to their commonsense score, and calculate the precision and recall curve by comparing the top predictions with held-out triplets. The AUC is computed by the area under curve. For each point on the curve, we compute the F1 score by the harmonic mean of precision and recall, and report the maximum F1 score on the curve. Precision@K\% is the precision of the top K\% extractions of the candidates. For macro evaluation, we first calculate the precision-recall curve of each relation, and then obtain the macro curve by the average of different relation curves.

\end{document}